\documentclass[
]{ceurart}

\usepackage{listings}
\usepackage{floatrow}
\usepackage{graphicx}
\usepackage{tabularx}
\usepackage{multirow}
\usepackage{subcaption}
\usepackage{xspace}

\begin{document}

\copyrightyear{2022}
\copyrightclause{Copyright for this paper by its authors.
  Use permitted under Creative Commons License Attribution 4.0
  International (CC BY 4.0).}

\conference{International Workshop on Data-driven Resilience Research 2022,
  July 06, 2022, Leipzig, Germany}

\title{Assessing Anonymized System Logs Usefulness for Behavioral Analysis in RNN Models}


\author[1]{Tom Richard Vargis}[%
email={tom_richard.vargis@tu-dresden.de},
]
\cormark[1]
\address[1]{Technische Universit\"at Dresden, Germany}
\address[2]{ScaDS.AI Dresden/Leipzig, Germany}

\author[1,2]{Siavash Ghiasvand}[%
email=siavash.ghiasvand@tu-dresden.de,
]

\cortext[1]{Corresponding author.}

\begin{abstract}
  System logs are a common source of monitoring data for analyzing computing systems behavior. Due to the complexity of modern computing systems and the large size of collected monitoring data, automated analysis mechanisms are required. Numerous machine learning and deep learning methods are proposed to address this challenge. However, due to the existence of sensitive data in system logs their analysis and storage raise serious privacy concerns. Anonymization methods could be used to cleanse the monitoring data before analysis. However, anonymized system logs in general do not provide an adequate usefulness for majority of behavioral analysis. Content-aware anonymization mechanisms such as $P\alpha RS$ preserve the correlation of system logs even after anonymization. This work evaluates the usefulness of anonymized system logs of Taurus HPC cluster anonymized using $P\alpha RS$, for behavioural analysis via recurrent neural network models. To facilitate the reproducibility and further development of this work, the implemented prototype and monitoring data are publicly available~\cite{ref_gitcode}.
\end{abstract}


\maketitle

\section{Introduction}
\label{sec-motivation}
It is of great interest to monitor large computing systems’ behavior. This knowledge helps to improve availability, reduce further damages caused by detectable failures, and diagnose problems. Despite the independent functionality of computing nodes in large computing systems, their behavior is highly dependent on other components of the computing system. The hierarchical design (e.g., Fat tree topology) of large computing systems, such as high-performance clusters (HPC), and the utilization of shared resources in such systems are the main reason for behavioral dependency among the computing nodes. Furthermore, the strategies employed to determine the usage of node sets to further enhance system performance (e.g., utilizing neighboring nodes) also have a direct impact on behavioral dependencies among computing nodes. Earlier studies identified strong spatial and temporal correlations among computing nodes of large computing systems~\cite{ref_lessons2016}.

In recent years, numerous automatic and semi-automatic behavioral analysis methods have been proposed.
These methods utilize various monitoring data such as data collected by hardware sensors, system logs, batch system information, and user activity logs to detect and predict the system behavior.
Due to the complexity of large-scale computing systems and their dynamic nature, identifying a system's behavioral pattern is challenging.
Furthermore, sensitive information contained in the monitoring data (e.g., user activity logs), has raised privacy concerns about the use of some analytical methods as well as the outsourcing of analysis.

The anonymization method P$\alpha$RS~\cite{ref_ficc2018} has been proposed to address the privacy concerns of processing monitoring data containing sensitive information.
Preliminary results indicate the usefulness of such anonymized system logs for the detection of abnormal behaviors in HPC systems via auto-encoders~\cite{ref_icpram2019}.
This study examines the effectiveness of using fully anonymized system logs in one of the most commonly used models of recurrent neural networks (RNN) for anomaly detection, namely Long short-term memory or LSTM.
Due to the nature of these analyses, a short return time, as well as a short training time, is required.
Therefore, the model needs to be kept as simple as possible, and it should be possible to train the model with as little data as possible.

The remainder of this work is structured as follows: Section~\ref{sec-related-work} provides an overview of using deep learning methods for behavioral analysis, with a focus on LSTM. Section~\ref{sec-materials-and-methods} describes the monitoring data, the preprocessing steps, and the main parameters used in this work.
The fitness of the proposed model for this work is verified in Section~\ref{subsec-lstm-model}.
In Section~\ref{sec-result} the results of each experiment have been discussed. Finally, Section~\ref{sec-conclusion} concludes the work and specifies the important future work directions.

\section{Related works}
\label{sec-related-work}
Anomaly detection is a pivotal part of system log analysis and has been the subject of numerous types of research. Among common deep learning models for anomaly detection, LSTM has been widely employed due to its success in providing highly accurate predictions. Qicheng Ma et al. in~\cite{ref_DANTE2021} and Min Du et al. in~\cite{ref_deeplog} modeled system logs as natural language sequences and patterns were extracted from these sequences. This analysis was done to detect insider threats and any deviations from these sequences were seen as a potential threat.
A similar approach was used in~\cite{ref_NlpDL2018} which had a feature extraction algorithm like Word2vec and then employed an LSTM for anomaly detection.

In \cite{ref_automatedIT2016} log patterns from heterogenous logs were extracted by clustering similar logs together and from these patterns, sequential features over time were extracted.
These features over time were finally passed through LSTM to detect failures.
Log parsing and feature extraction followed by two LSTMs and an Autoencoder was introduced in \cite{ref_Experiencereport2022} for failure detection.
An abnormal instance usually manifests itself as an outlier that significantly deviates from such patterns.
Zhuangbin et. al. concluded that log semantics indeed improves models’ robustness against noises.

Hao Chen et al. in~\cite{ref_unsupervised} proposed a novel semantic information embedding technique to detect anomalies. Some keywords in log entries may represent the meaning of the entire system logs.
A CNN combined with the LSTM approach not only learns semantics but also the quantitative feature from the log count vector.   
A slightly different approach is proposed in~\cite{ref_systemfailure}.
Yixin et. al. in addition to the deep learning model takes a step further by closely examining the timestamps of the log data which majority of existing studies have generally ignored.
Yixin et. al. propose to integrate log timestamps in deep learning models using interpolation techniques.
This addition was proved to improve the ultimate accuracy of failure detection.

The above-mentioned studies deliver a detection accuracy of greater than 93\%\footnote{In controlled environment and with adequate data preparation steps, near to perfect accuracy is possible.}.
Whilst some of these approaches use a supervised method where predefined ground truth is set as the pattern and any deviation from these were classified as an anomaly, some others use an unsupervised approach where patterns were identified from the extracted sequential features by the monitoring data over time.
The common point in all the above studies is the usage of monitoring data in its original format.
The existence of sensitive information in monitoring data raises serious concerns in many use cases, the storage of monitoring data becomes challenging and the outsourcing of analysis is not possible.

Conversely, to the aforementioned studies, the approach proposed in this work employs fully anonymized system logs.
Thus, eliminating all privacy concerns and making it possible to outsource the entire log analysis process.
On the other hand, the usage of anonymized monitoring data makes the identification process increasingly challenging since the content of the encoded logs cannot be retrieved.


\section{Method and Data}
\label{sec-materials-and-methods}
Recurrent Neural Networks (RNN) are known to perform well on time series data.
The model, Long Short-Term Memory (LSTM) is a special type of RNN capable of learning dependencies between the data and making sequential predictions.
This work builds an LSTM model that requires short memory of the past to identify and predict the pattern of upcoming log messages.
The log messages are anonymized in a pre-processing step based on the P$\alpha$RS mechanism~\cite{ref_ficc2018}.
This is achieved by classifying messages with similar patterns into a single class (pattern) and then generating a unique hash key for each pattern.
Both univariate and multivariate data were tested on this model.
The final goal is to assess the usefulness of anonymized system logs for anomaly detection via LSTM models.
Thus, it is essential to avoid unnecessary complexities.
To achieve this goal, a one-layer LSTM model followed by one dense layer was selected.
The Keras\footnote{Available at https://keras.io/about/.} library was used to implement the model.

In this work, Adam is used as the optimizer and the Mean Absolute Error (MAE)is employed as the loss function.
Mean Squared Logarithmic Error (MSLE) is also used in the later parts of the report.
MSLE considers the relative difference between the real and the predicted value.
As the data used for the analysis is normalized, choosing MAE over MSLE is not expected to make a notable difference to the error values.
In the testing dataset, a point is classified as an anomaly if the MAE loss goes beyond the specified threshold. Here the threshold is defined as the maximum value of the MAE loss for the training dataset.

System Logging Protocol or Syslog is the common protocol used to send system logs or event messages to a specific server, called a Syslog server.
It is primarily used to collect various device logs from several machines in a central location for monitoring and review.
Syslog is available in all Unix and Linux-based systems.
As all the TOP500\footnote{https://top500.org} HPC systems are Linux-based (at the time of writing), Syslog analysis applies to all HPC systems.
The detailed specification of the Syslog protocol is defined in RFC5424\footnote{https://datatracker.ietf.org/doc/html/rfc5424}.

Taurus\footnote{Detailed hardware information: https://tud.link/7y2h} HPC cluster, located in Dresden, has around 2000 compute nodes including 750 GPUs and a total count of 80,000 CPU cores.
Taurus is divided into several sections known as Islands.
Each Island has its specific hardware configuration.
Island 8 of Taurus is powered by AMD Rome CPUs and NVIDIA A100 GPUs.
Thus, one of the most utilized islands on Taurus by numerous users and for various applications.
For this work, the system logs of first the 16 nodes of Island 8 are used as the source of monitoring data.
This selection is based on the idea that Island 8 of Taurus is an active Island and the first 16 nodes are known to have shared resources.

For the multivariate model, four features were considered in a specified time bucket (eg: 10 min), namely the average severity, average facility, the frequency of top 10 messages\footnote{The top 10 classes of syslog messages with highest frequency.} from the last 24 hours, and the frequency of non-top 10 messages from the last 24 hours.

Furthermore, for the univariate model two synthesized datasets, one with repetitive patterns and the other with significant noises were examined.
The goal of using synthesized datasets was to identify the similarities in the model's behavior based on the input data.

The model and the collected monitoring data have several adjustable parameters which can be fine-tuned to improve the prediction accuracy.
Table~\ref{tab_hyper_parameters} provides a list of parameters used in this work.

\begin{table}[ht]
	\caption{List of metrics and hyper-parameters}
	\label{tab_hyper_parameters}
	\centering
	\begin{tabular}{|l|l|}
		\hline
		\textbf{Parameter}     				      & \textbf{Possible values}                  \\
		\hline
		Stateful                          & Fixed                   \\
		Node bucket                       & Fixed (16 nodes Island8) \\
		Learning rate                     & 0.001, 0.01, 0.1        \\
		Epochs                            & 20, 50, 100, 10000      \\
		Number of steps                   & 2, 4, 6, 12                \\
		Time bucket                       & 5 min, 10 min, 30 min   \\
		Cumulative Sum                    & Yes, No                 \\
		Normalization                     & None, MinMax, Sigmoid   \\
		Number of top messages considered & Top3, Top5, Top10   	\\
		\hline
	\end{tabular}
\end{table}

The \textit{learning rate} hyperparameter used in the training of the LSTM is given a value between 0.0 to 1.0.
Four different learning rates [0.1,0.01,0.001,0.0001] were used in the first experiments.
However, to better observe the relevance and impact of learning rate\footnote{From various studies it is known that large values of learning rate may impose destructive impact on the learning process.}, the three larger rates were used in most experiments.
Although in the Keras\footnote{Available from https://keras.io/.} framework, by default, LSTMs are stateless, to consider dependencies between batches, the model needed to be stateful.
The \textit{time bucket} shows the time interval in which the monitoring data is aggregated.
For practical reasons detecting anomalies with a delay of more than 30 minutes is not favorable.
The \textit{number of steps} in principle is the memory of the LSTM.
The model uses this number of steps to predict the following step.
For example, if the \textit{time bucket} is given as 10 minutes and the number of steps as 6, then 60 minutes of data is used to predict the next data point.
\textit{Node bucket} defines the number of nodes selected from an Island on Taurus.
Considering the significant similarity in behavior of the neighboring nodes\footnote{also known as node vicinity~\cite{ref_ispdc2019}} due to shared resources, the first 16 nodes of Island 8 are taken for training the model.

\textit{Cumulative sum} of the count of messages is an additional setup where the frequency features are cumulatively summed up every hour.
This accumulation amplifies the system logs' hourly pattern facilitating identification of the same by the model.
Data is \textit{normalized} either using the MinMax scaler function which normalizes the dataframe according to the maximum value or via a sigmoid function which adjusts the data to the scale of 0 to 1.

\section{Model fitness}
\label{subsec-lstm-model}

\begin{figure}[!htb]
	\begin{subfigure}[b]{0.32\textwidth}
		\includegraphics[width=1\textwidth]{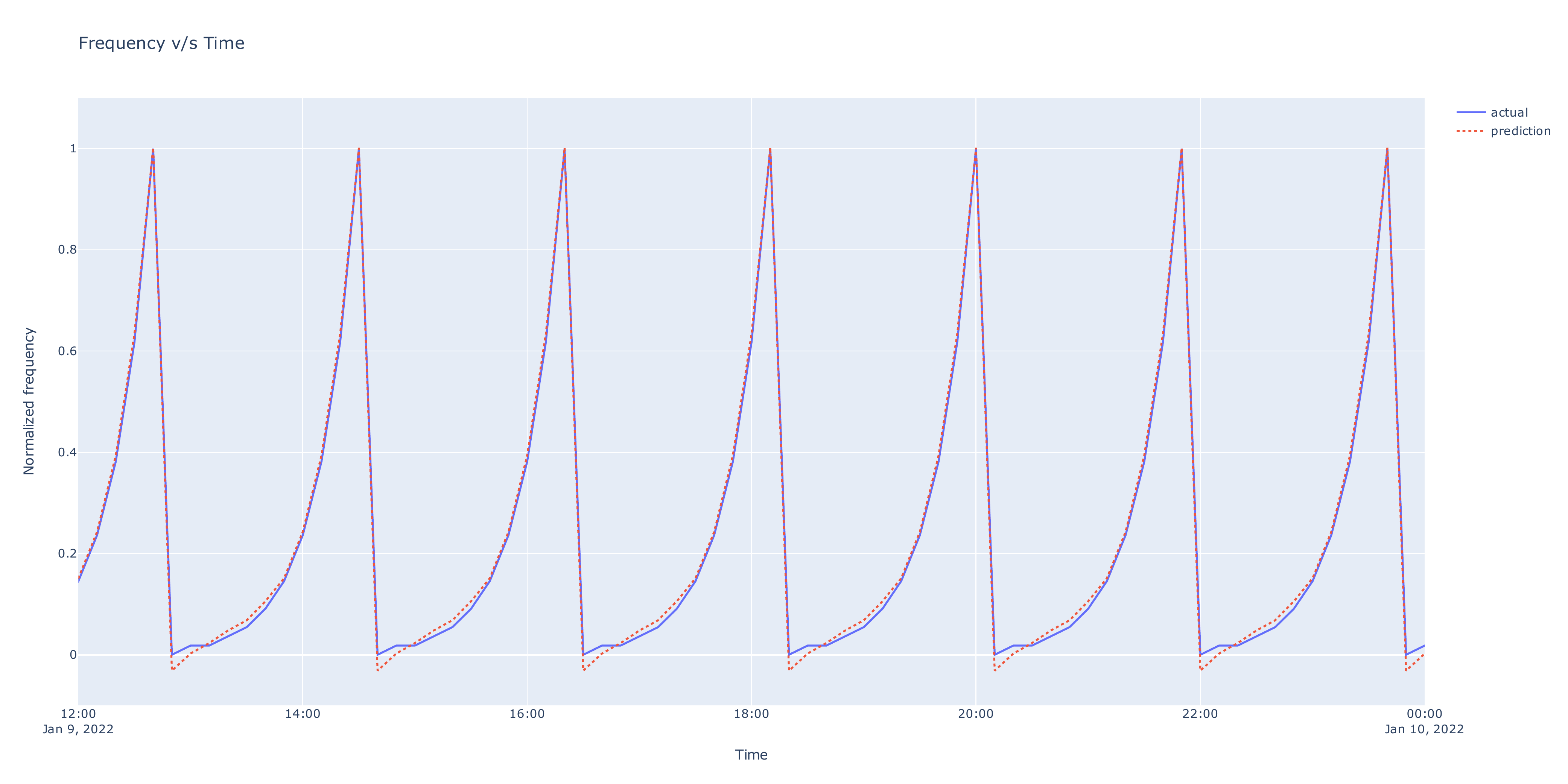}
		\caption{Fibonacci dataset}
		\label{fig-freq-vs-time-plot12}
	\end{subfigure}
	\begin{subfigure}[b]{0.32\textwidth}
		\includegraphics[width=1\textwidth]{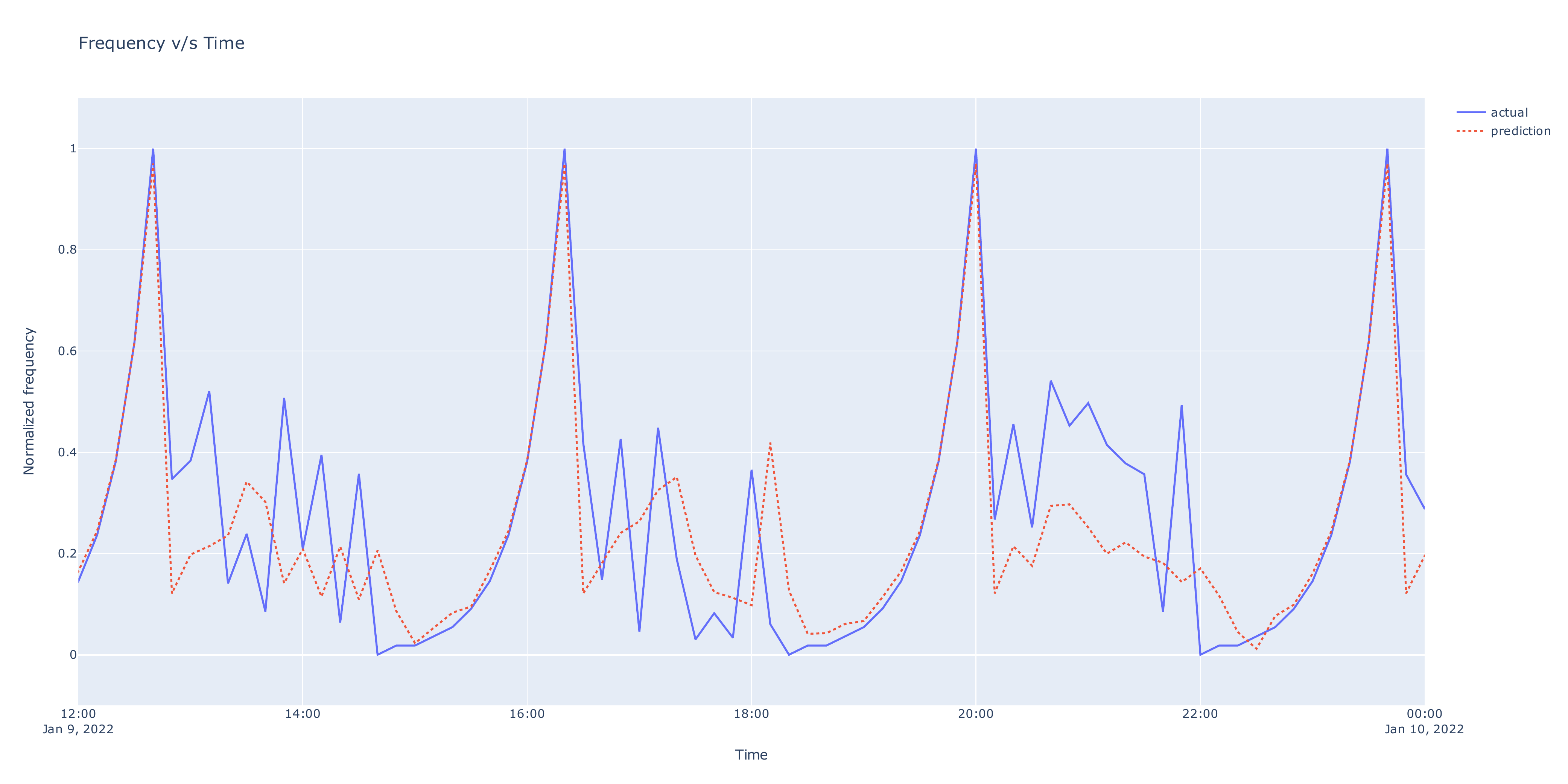}
		\caption{Partially randomized}
		\label{fig-freq-vs-time-plot13}
	\end{subfigure}
	\begin{subfigure}[b]{0.32\textwidth}
		\includegraphics[width=1\textwidth]{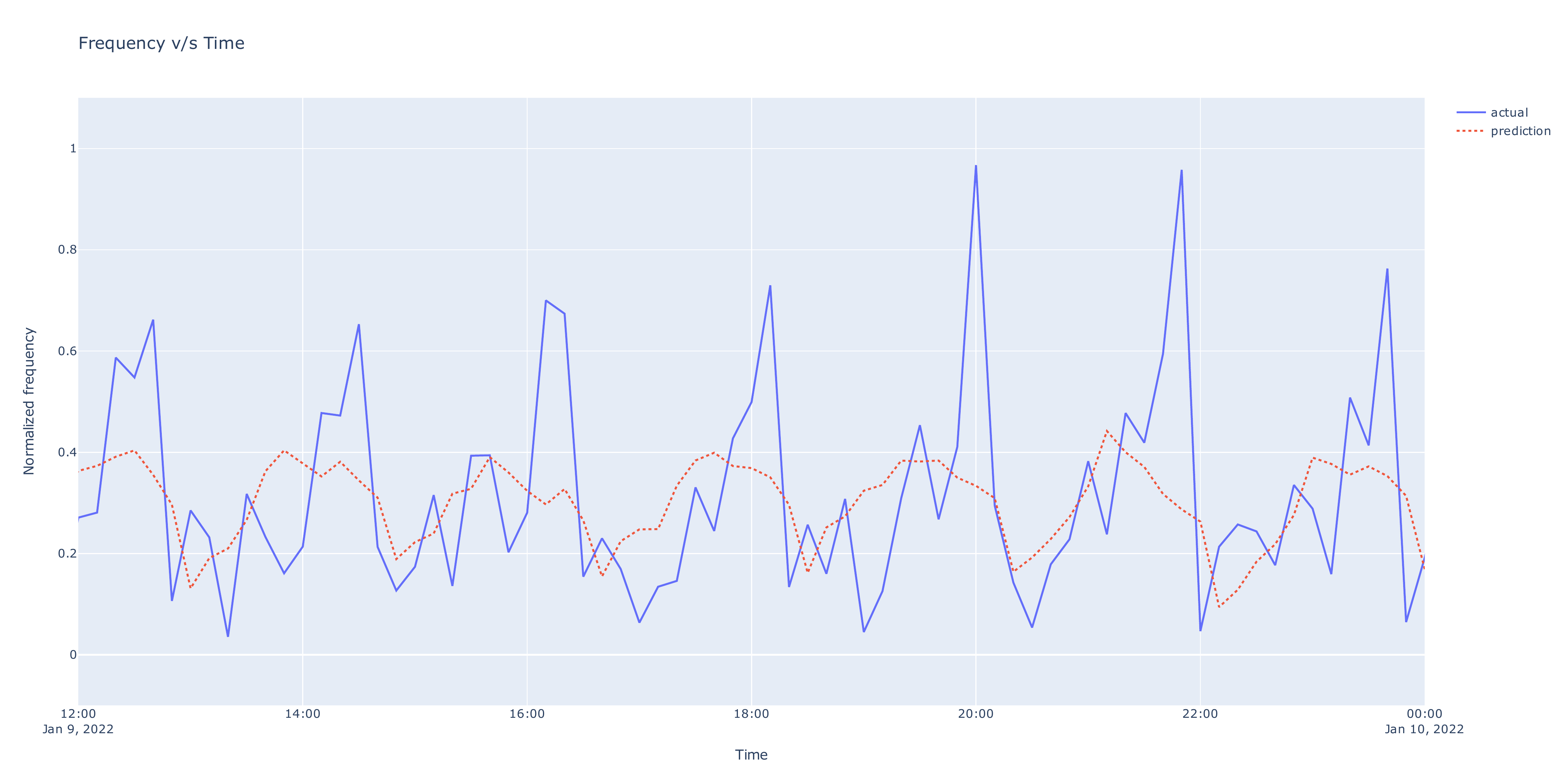}
		\caption{Detectable trends}
		\label{fig-freq-vs-time-plot14}
	\end{subfigure}
	\caption{Assessment of LSTM model using synthesized dataset}
	\label{fig-freq-vs-time-plot13-14}
\end{figure}

To evaluate the fitness of the proposed model for the intended purpose of this work, a synthesized dataset with repetitive patterns was constructed.
The synthesized dataset consisted of the first 10 Fibonacci numbers repeated 100 times.
The proposed model was able to identify and learn the repetitive pattern and predict the series with high accuracy as shown in Figure~\ref{fig-freq-vs-time-plot12}.

Furthermore, this data was slightly altered by having the first 10 Fibonacci numbers followed by 10 random numbers from the range of 0 to 30.
This set of 20 numbers was then repeated 100 times, to form the test dataset.
Figure~\ref{fig-freq-vs-time-plot13} shows that the model successfully recognized the repetitive pattern of the Fibonacci part and resembled the random part of the data with a short delay.

Finally, a random noise (in the range of 0 to 30) was added to the initial dataset and the same model was trained using this new dataset.
This time, as shown in Figure~\ref{fig-freq-vs-time-plot14} the predictions lag and, as expected, the model could only learn the general trend in data.

From these observations, it is concluded that the proposed model can learn patterns and trends in data despite the small size of the dataset, and the low number of epochs.
It is important to note that most applications of log analysis (e.g. live anomaly detection) require a short execution time.
Therefore, using small datasets and a low number of epochs is highly favorable.

\section{Results and Discussions}
\label{sec-result}
Different combinations of parameters were tested for both univariate and multivariate datasets to evaluate their impact on the model's prediction.
The results of both uni- and multivariate datasets are shown in Table~\ref{tab-training-validation-uni-multi}.

\begin{table}[]
	\caption{Comparison of loss values for varying learning rates and number of steps}
	\label{tab-training-validation-uni-multi}
	\resizebox{1\textwidth}{!}
	{
		\small
		\begin{tabular}{lll|llll|llll|}
			\multicolumn{11}{c}{Univariate}\\
			\hline
			\multicolumn{3}{|l|}{Time bucket} &
			\multicolumn{4}{l|}{10-min} &
			\multicolumn{4}{l|}{30-min} \\ \hline
			\multicolumn{3}{|l|}{Number of steps} &
			\multicolumn{1}{l|}{2} &
			\multicolumn{1}{l|}{6} &
			\multicolumn{1}{l|}{9} &
			12 &
			\multicolumn{1}{l|}{2} &
			\multicolumn{1}{l|}{6} &
			\multicolumn{1}{l|}{9} &
			12 \\ \hline
			\multicolumn{3}{|l|}{Batch size} &
			\multicolumn{1}{l|}{134} &
			\multicolumn{1}{l|}{78} &
			\multicolumn{1}{l|}{89} &
			622 &
			\multicolumn{1}{l|}{125} &
			\multicolumn{1}{l|}{27} &
			\multicolumn{1}{l|}{206} &
			123 \\ \cline{1-3}
			&
			\multicolumn{1}{l|}{} &
			Lr &
			\multicolumn{1}{l|}{} &
			\multicolumn{1}{l|}{} &
			\multicolumn{1}{l|}{} &
			&
			\multicolumn{1}{l|}{} &
			\multicolumn{1}{l|}{} &
			\multicolumn{1}{l|}{} &
			\\ \hline
			\multicolumn{1}{|l|}{\multirow{6}{*}{\begin{tabular}[c]{@{}l@{}}Loss\\ value\end{tabular}}} &
			\multicolumn{1}{l|}{Training} &
			\multirow{2}{*}{0.001} &
			\multicolumn{1}{l|}{0.050} &
			\multicolumn{1}{l|}{0.041} &
			\multicolumn{1}{l|}{0.047} &
			0.061 &
			\multicolumn{1}{l|}{0.074} &
			\multicolumn{1}{l|}{0.070} &
			\multicolumn{1}{l|}{0.068} &
			0.070 \\ \cline{2-2} \cline{4-11} 
			\multicolumn{1}{|l|}{} &
			\multicolumn{1}{l|}{Validation} &
			&
			\multicolumn{1}{l|}{0.057} &
			\multicolumn{1}{l|}{0.049} &
			\multicolumn{1}{l|}{0.052} &
			0.081 &
			\multicolumn{1}{l|}{0.088} &
			\multicolumn{1}{l|}{0.076} &
			\multicolumn{1}{l|}{0.091} &
			0.087 \\ \cline{2-11} 
			\multicolumn{1}{|l|}{} &
			\multicolumn{1}{l|}{Training} &
			\multirow{2}{*}{0.01} &
			\multicolumn{1}{l|}{0.051} &
			\multicolumn{1}{l|}{0.038} &
			\multicolumn{1}{l|}{0.033} &
			0.044 &
			\multicolumn{1}{l|}{0.069} &
			\multicolumn{1}{l|}{0.055} &
			\multicolumn{1}{l|}{0.058} &
			0.058 \\ \cline{2-2} \cline{4-11} 
			\multicolumn{1}{|l|}{} &
			\multicolumn{1}{l|}{Validation} &
			&
			\multicolumn{1}{l|}{0.057} &
			\multicolumn{1}{l|}{0.042} &
			\multicolumn{1}{l|}{0.037} &
			0.058 &
			\multicolumn{1}{l|}{0.083} &
			\multicolumn{1}{l|}{0.065} &
			\multicolumn{1}{l|}{0.079} &
			0.066 \\ \cline{2-11} 
			\multicolumn{1}{|l|}{} &
			\multicolumn{1}{l|}{Training} &
			\multirow{2}{*}{0.1} &
			\multicolumn{1}{l|}{0.066} &
			\multicolumn{1}{l|}{0.218} &
			\multicolumn{1}{l|}{0.603} &
			0.082 &
			\multicolumn{1}{l|}{0.068} &
			\multicolumn{1}{l|}{0.506} &
			\multicolumn{1}{l|}{1.846} &
			0.082 \\ \cline{2-2} \cline{4-11} 
			\multicolumn{1}{|l|}{} &
			\multicolumn{1}{l|}{Validation} &
			&
			\multicolumn{1}{l|}{0.085} &
			\multicolumn{1}{l|}{0.314} &
			\multicolumn{1}{l|}{0.693} &
			0.100 &
			\multicolumn{1}{l|}{0.081} &
			\multicolumn{1}{l|}{0.796} &
			\multicolumn{1}{l|}{2.159} &
			0.099 \\ \hline
		\end{tabular}
		
		\begin{tabular}{lll|ll|ll|}
			\multicolumn{7}{c}{Multivariate}\\
			\hline
			\multicolumn{3}{|l|}{Epochs}                                                      & \multicolumn{2}{l|}{50}              & \multicolumn{2}{l|}{100}             \\ \hline
			\multicolumn{3}{|l|}{Number of steps}                                             & \multicolumn{1}{l|}{2}      & 6      & \multicolumn{1}{l|}{2}      & 6      \\ \hline
			\multicolumn{3}{|l|}{Batch size}                                                  & \multicolumn{1}{l|}{199}    & 1      & \multicolumn{1}{l|}{125}    & 27     \\ \cline{1-3}
			& \multicolumn{1}{l|}{}           & Lr          & \multicolumn{1}{l|}{}       &        & \multicolumn{1}{l|}{}       &        \\ \hline
			\multicolumn{1}{|l|}{\multirow{6}{*}{\begin{tabular}[c]{@{}l@{}}Loss\\ value\end{tabular}}} &
			\multicolumn{1}{l|}{Train.} &
			\multirow{2}{*}{0.0001} &
			\multicolumn{1}{l|}{0.147} &
			0.124 &
			\multicolumn{1}{l|}{0.145} &
			0.121 \\ \cline{2-2} \cline{4-7} 
			\multicolumn{1}{|l|}{} & \multicolumn{1}{l|}{Val.} &                        & \multicolumn{1}{l|}{0.114} & 0.077 & \multicolumn{1}{l|}{0.113} & 0.079 \\ \cline{2-7} 
			\multicolumn{1}{|l|}{} & \multicolumn{1}{l|}{Train.}   & \multirow{2}{*}{0.001} & \multicolumn{1}{l|}{0.149} & 0.113  & \multicolumn{1}{l|}{0.139} & 0.092 \\ \cline{2-2} \cline{4-7} 
			\multicolumn{1}{|l|}{} & \multicolumn{1}{l|}{Val.} &                        & \multicolumn{1}{l|}{0.112} & 0.074 & \multicolumn{1}{l|}{0.109} & 0.080   \\ \cline{2-7} 
			\multicolumn{1}{|l|}{} & \multicolumn{1}{l|}{Train.}   & \multirow{2}{*}{0.01}  & \multicolumn{1}{l|}{0.140} & 0.114 & \multicolumn{1}{l|}{0.145}  & 0.115 \\ \cline{2-2} \cline{4-7} 
			\multicolumn{1}{|l|}{} & \multicolumn{1}{l|}{Val.} &                        & \multicolumn{1}{l|}{0.105} & 0.085 & \multicolumn{1}{l|}{0.108} & 0.078 \\ \hline
		\end{tabular}
	}
\end{table}

For the univariate dataset, in each experiment, one parameter takes different values from the possible values, while the other parameters remain constant.
The set of parameters used in each experiment is shown as a tuple.
The "*" represents the varying parameter.

\begin{figure}[!htb]
	\begin{subfigure}[b]{0.49\textwidth}
		\includegraphics[width=1\textwidth]{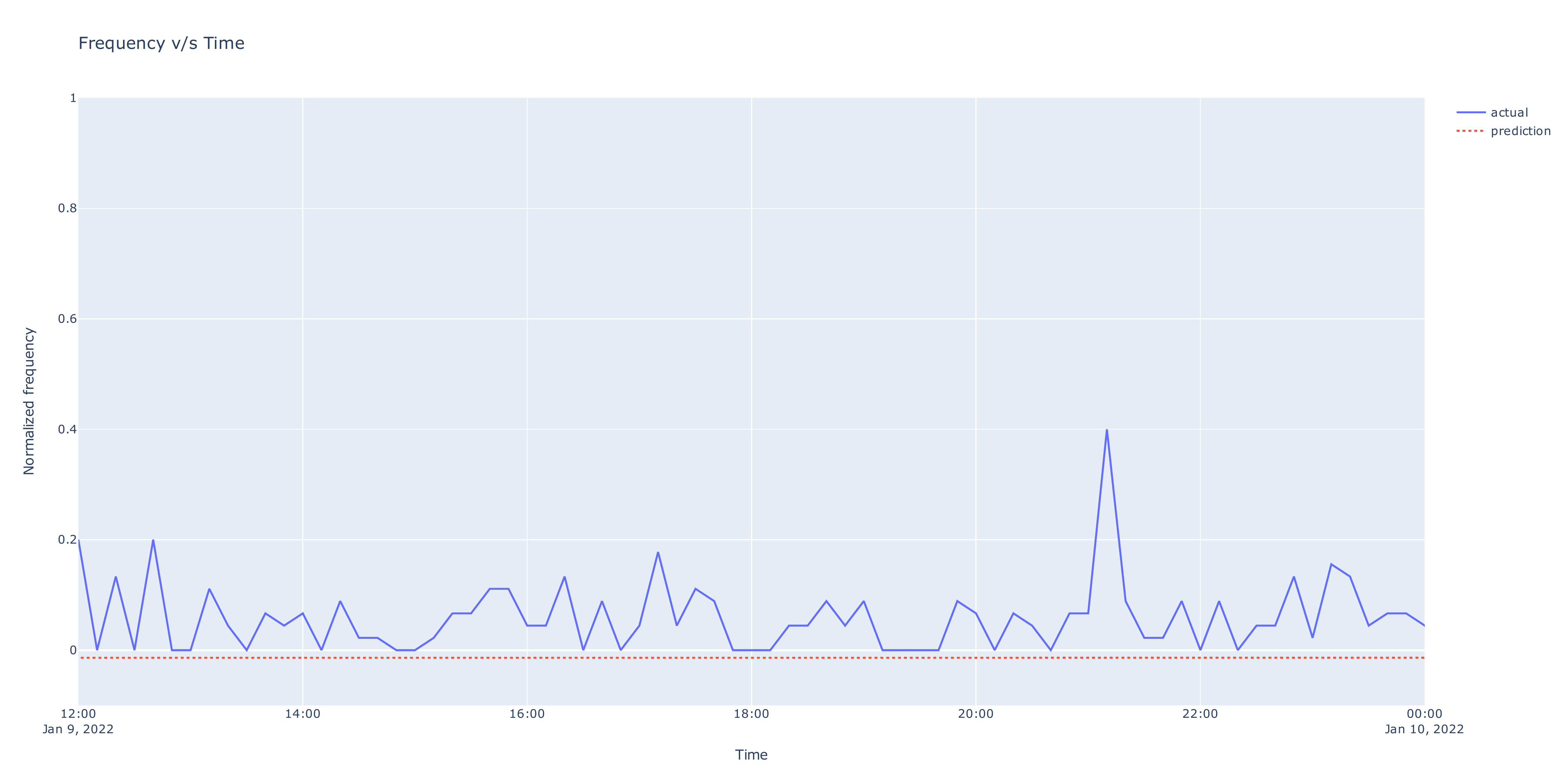}
		\caption{Lr = 0.1}
		\label{fig-freq-vs-time-plot2-a}
	\end{subfigure}
	\begin{subfigure}[b]{0.49\textwidth}
		\includegraphics[width=1\textwidth]{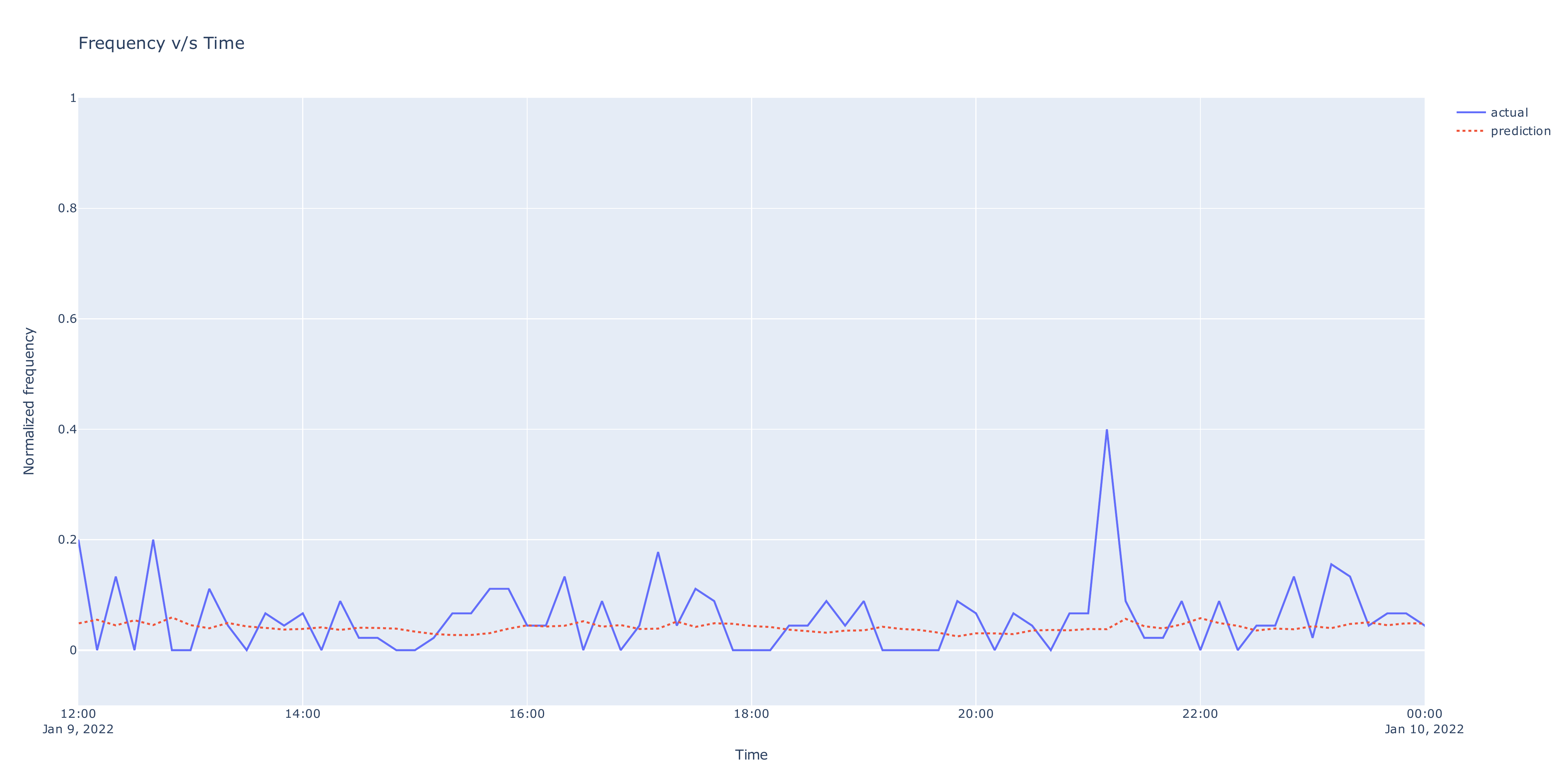}
		\caption{Lr = 0.001}
		\label{fig-freq-vs-time-plot2-b}
	\end{subfigure}
	\caption{Impact of learning rate on predictions (time bucket = 10 min)}
	\label{fig-freq-vs-time-plot2}
\end{figure}

\paragraph{Learning rate (Lr)}
$[ *, 100, 6, 10 min, No, MinMax, top10]$\\
While at a learning rate of 0.1 the model fails as expected (Figure~\ref{fig-freq-vs-time-plot2-a}), it yields slightly better predictions for a learning rate of 0.001 (Figure~\ref{fig-freq-vs-time-plot2-b}).

\begin{figure}[!htb]
	\begin{subfigure}[b]{0.49\textwidth}
		\includegraphics[width=1\textwidth]{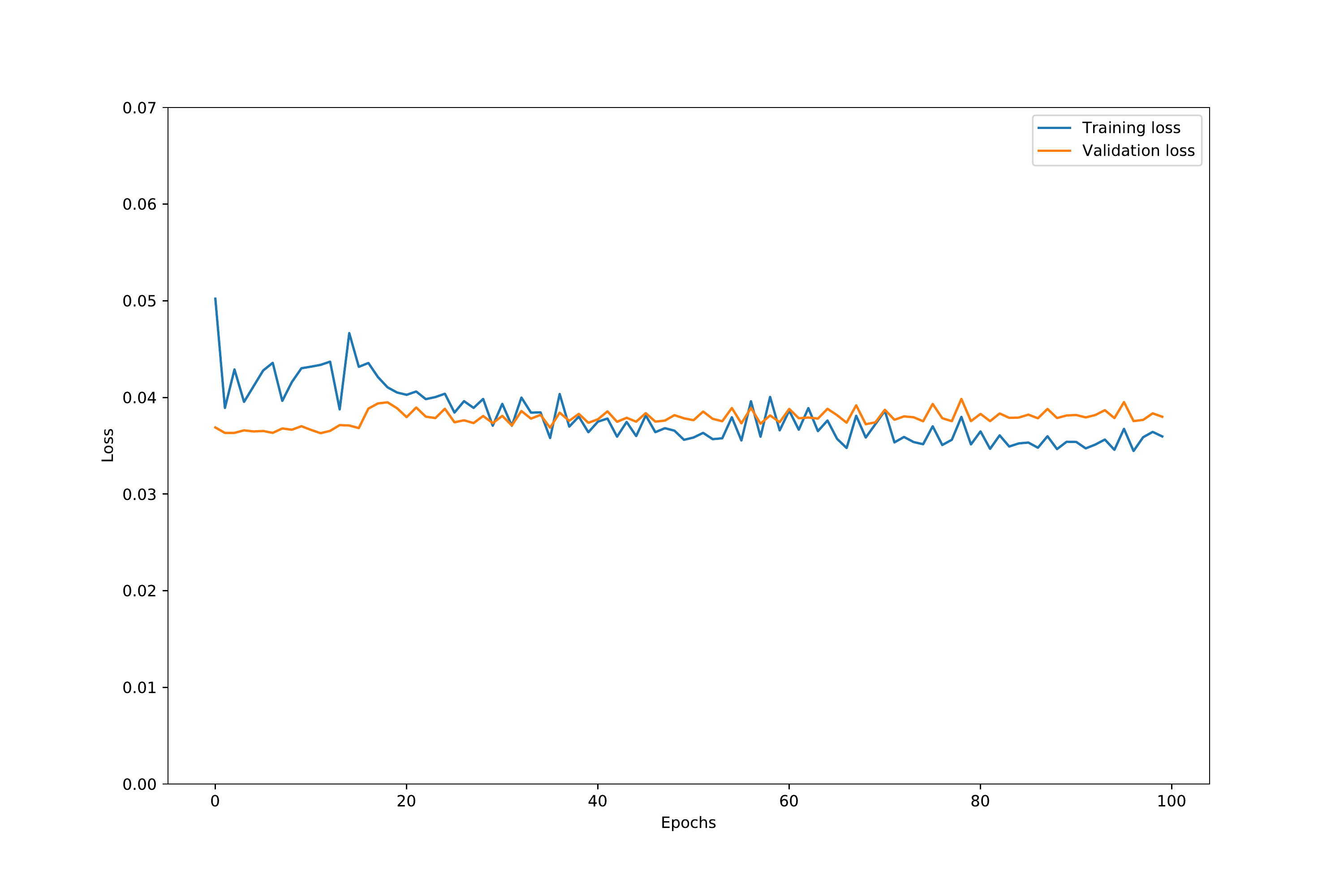}
		\caption{Epoch = 100}
		\label{fig-freq-vs-time-plot3-a}
	\end{subfigure}
	\begin{subfigure}[b]{0.49\textwidth}
		\includegraphics[width=1\textwidth]{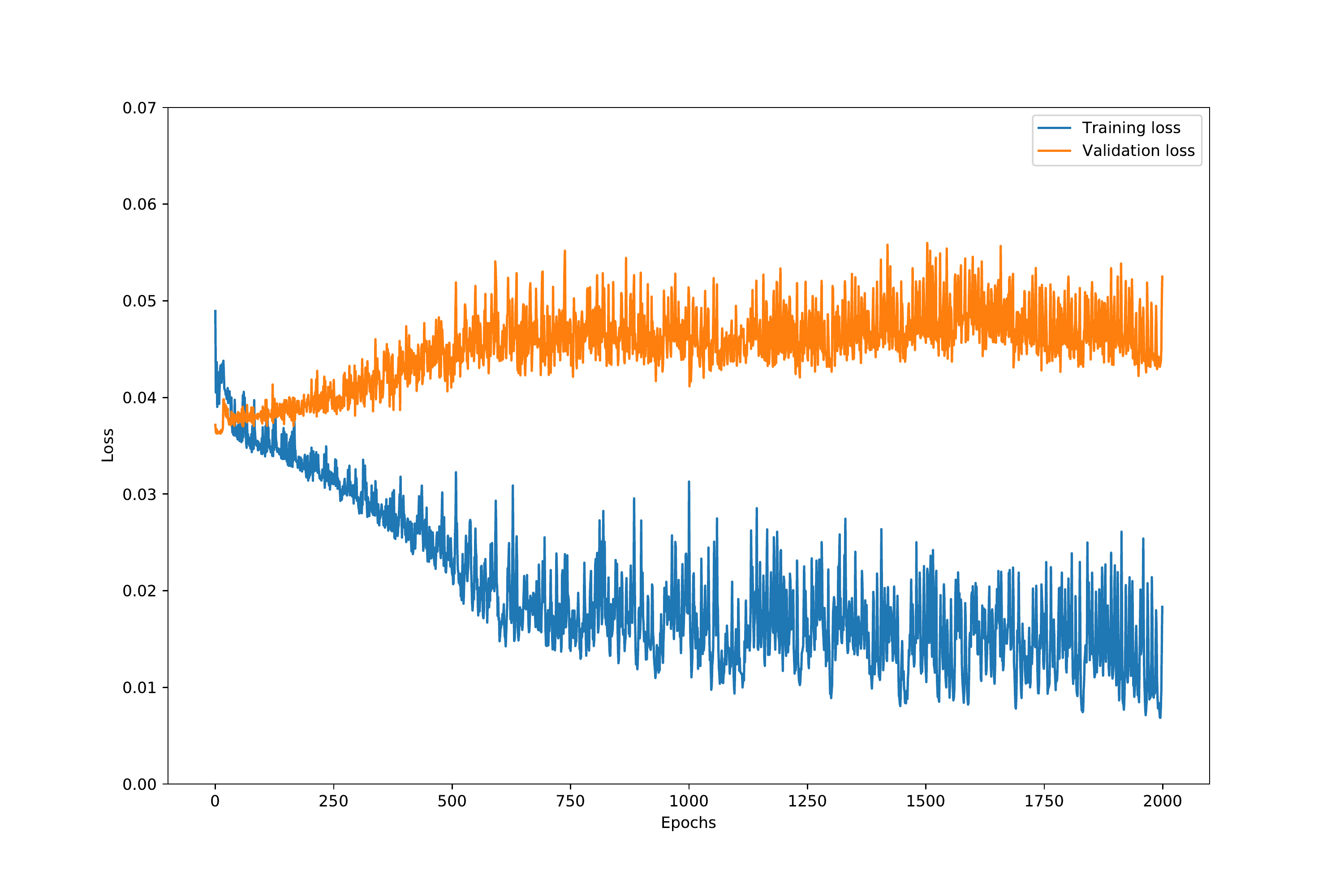}
		\caption{Epoch = 10,000}
		\label{fig-freq-vs-time-plot3-b}
	\end{subfigure}
	\caption{Impact of different amount of epochs on loss value}
	\label{fig-freq-vs-time-plot3}
\end{figure}

\paragraph{Epochs}
$[ 0.01, *, 6, 10 min, No, MinMax, top10]$\\
With a low number of epochs, e.g., 20 epochs, sometimes the error is not sufficiently small and with large epochs, e.g., 100 or 200 epochs, the error starts to oscillate towards the end.
There is no significant change observed in model behavior for epochs beyond 100.
A comparison between 100 and 10000 epochs is shown in Figure~\ref{fig-freq-vs-time-plot3}.
The overfitting problem with a large number of epochs is visible in Figure~\ref{fig-freq-vs-time-plot3-b}.
Similar observations were made in all experiments.
Therefore, the decision was made to fix the epochs at 50 so that the error converges.

\paragraph{Number of steps (Ns)}
$[ 0.01, 50, *, 10 min, No, MinMax, top10]$\\
Although the number of steps is an adjustable parameter it is also connected to the time bucket size.
Six steps for a 10 min time bucket means that the model uses 60 minutes of data (6 rows of data) to predict the next row.
Training on a few steps, e.g., 2 steps, reduces the model's memory.
Thus, instead of predicting future behavior, the model memorizes the patterns observed in recent time steps.
In contrast, using a large number of time steps, e.g., 12 steps, the model will be able to generalize and predict the overall trend in data.

\paragraph{Time bucket}
$[ 0.01, 50, 6, *, No, MinMax, top10]$\\
The data collected within a time bucket of 5 minutes contains too many null values leading to wrong predictions.
The 10 and 30-minute buckets provide relatively better data (less null values).
In addition, Taurus' behavior is highly dependent on the behavior of its users.
Hence, it shows hourly and daily patterns.

\paragraph{Cumulative Sum}
$[0.01, 50, 6, 30 min, *, MinMax, top10]$\\
There is a significant improvement in predictions when the cumulative sum is introduced in any setup.
This is possibly due to the hourly patterns (influenced by user activities) which are known to exist within Taurus log messages.

\paragraph{Normalization}
$[0.01,50, 6, 30 min, Yes, *, top10]$\\
Using non-normalized data versus MinMax scaled data, slightly changes the final output.
Sigmoid normalization does not improve the predictions hence, these are excluded.
Since an effective improvement cannot be detected, the data in this work is always normalized with the MinMax scaler.

\paragraph{Number of top messages considered}
$[0.01, 50, 6, 30 min, Yes,  MinMax, *]$\\
The choice of the number of top messages slightly influences the predictions.
Considering fewer top messages makes the predictions marginally better.
However, this improvement is not significant enough to vary this parameter.

\paragraph{Univariate data}
From the above observations the parameter list to be varied is narrowed down to the Learning rate, Time bucket, and Number of steps.
Both time buckets of 10 minutes and 30 minutes are considered.
The learning rate and the number of steps are varied among the selected range.
The batch size is automatically selected by the code so that the model remains stateful.

\begin{figure}[!htb]
	\begin{subfigure}[t]{0.49\textwidth}
		\includegraphics[width=1\textwidth]{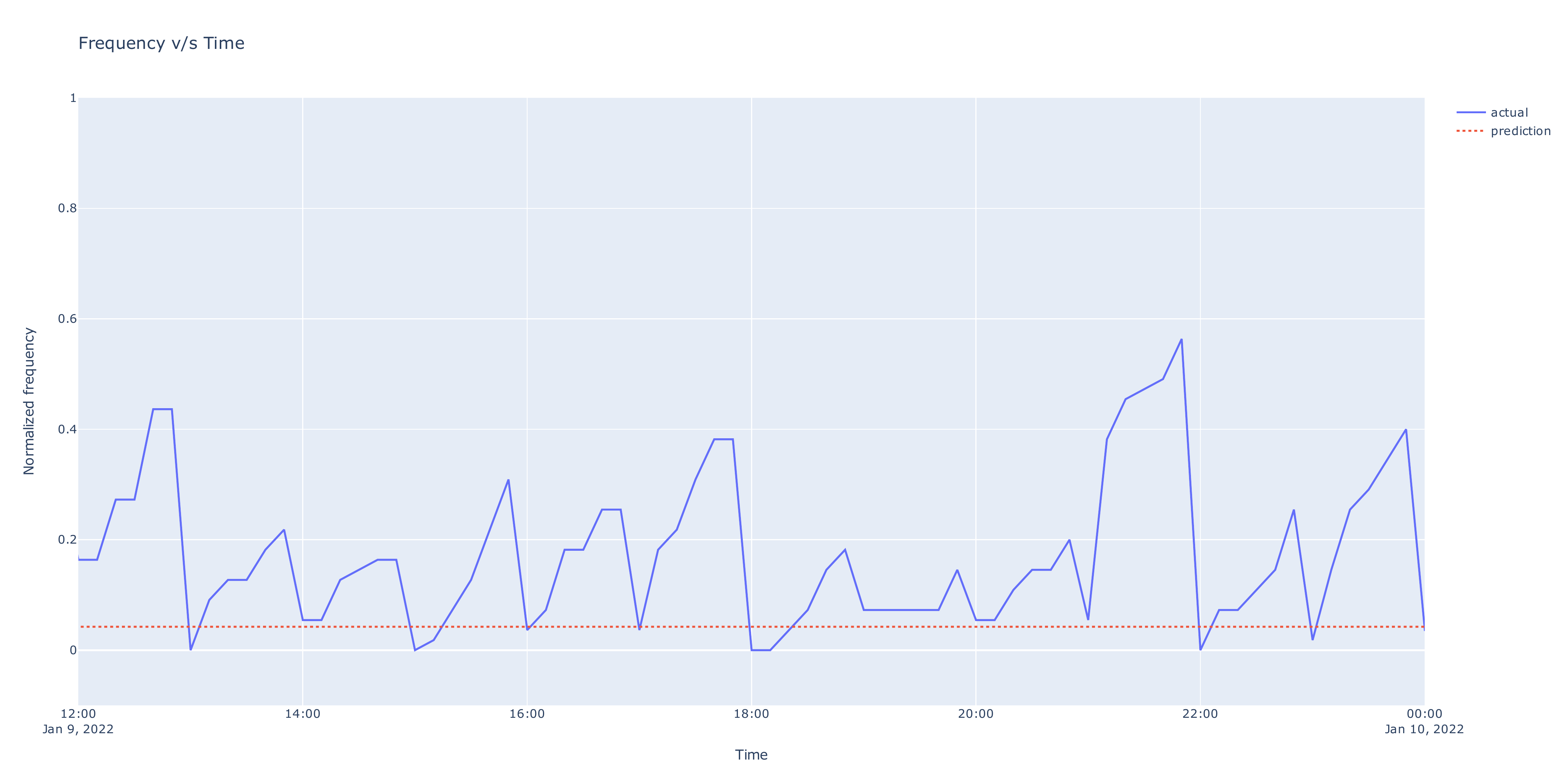}
		\caption{Lr = 0.1, Ns=6}
		\label{fig-freq-vs-time-plot6}
	\end{subfigure}
	\begin{subfigure}[t]{0.49\textwidth}
		\includegraphics[width=1\textwidth]{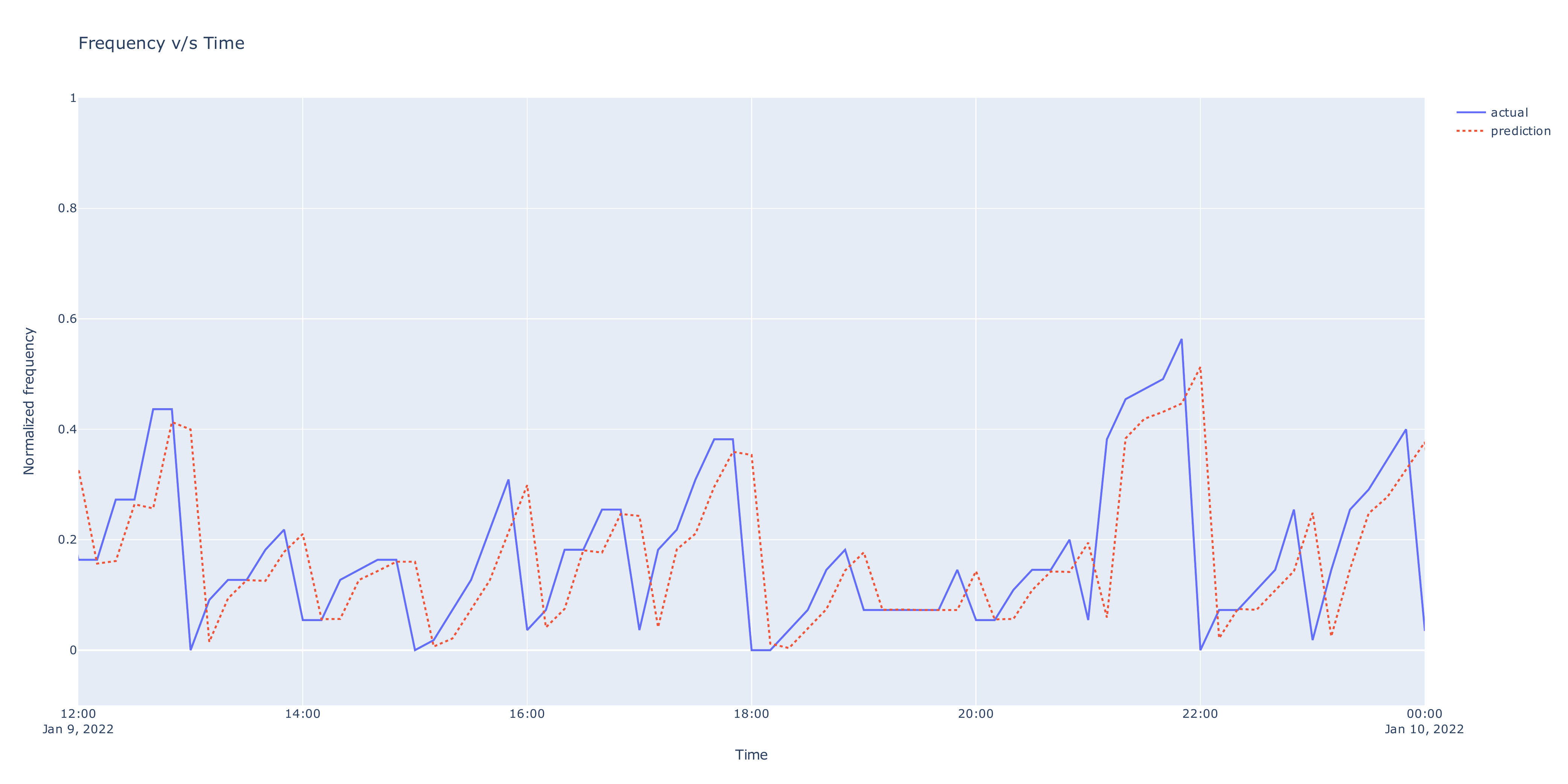}
		\caption{ Lr=0.001, Ns=2}
		\label{fig-freq-vs-time-plot5}
	\end{subfigure}
	\caption{Importance of learning rate and number of steps (time bucket = 10min)}
	\label{fig-freq-vs-time-plot5-6}
\end{figure}

The first observation made based on data collected in Table~\ref{tab-training-validation-uni-multi} is the role of different learning rates in prediction accuracy.
Regardless of the number of steps and the size of the time bucket, using a high value of learning rate ($0.1$), as shown in Fig.~\ref{fig-freq-vs-time-plot6} the model fails to learn the pattern of events as expected.
A high learning rate causes the model to take larger steps during the learning phase thus, it might converge to a suboptimal solution much quicker.

The learning rate of $0.001$, on the other hand, slows down the learning process.
Thus, with the given number of epochs, the model fails to make any meaningful predictions based on the learned information.
For the learning rate of $0.001$, the model is seen to have very limited memory.
Thus only replicates the very recent steps as shown in Fig.~\ref{fig-freq-vs-time-plot5}.
Although this behavior diminishes slightly as more time steps are considered, it does not improve the model to a noticeable extent.
For any choice of the number of steps, a learning rate of $0.01$ gives fairly good results among the three choices of learning rates.

\begin{figure}[!htb]
	\begin{subfigure}[t]{0.49\textwidth}
		\includegraphics[width=1\textwidth]{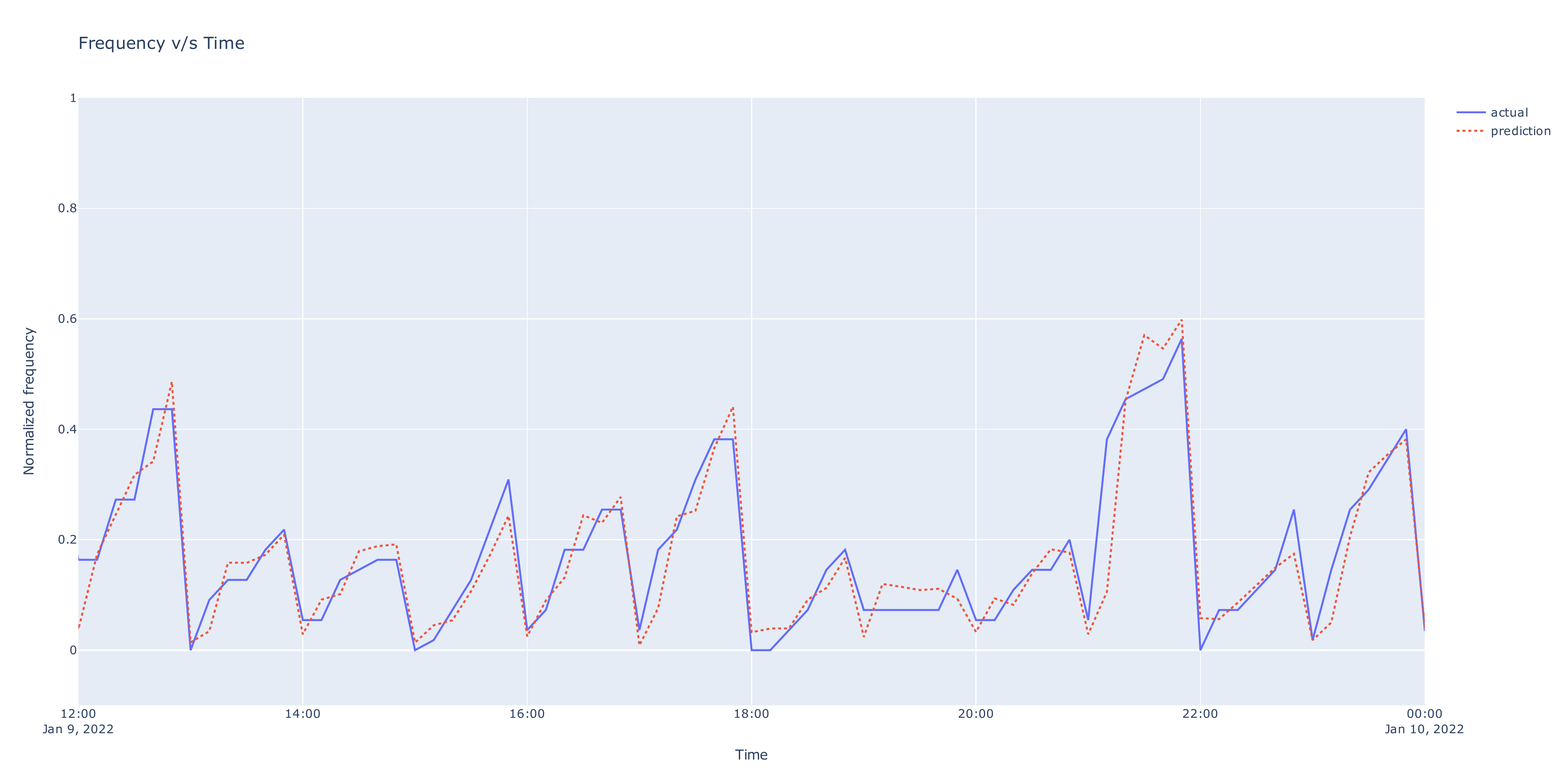}
		\caption{Lr=0.01, Ns=9}
		\label{fig-freq-vs-time-plot7}
	\end{subfigure}
	\begin{subfigure}[t]{0.49\textwidth}
		\includegraphics[width=1\textwidth]{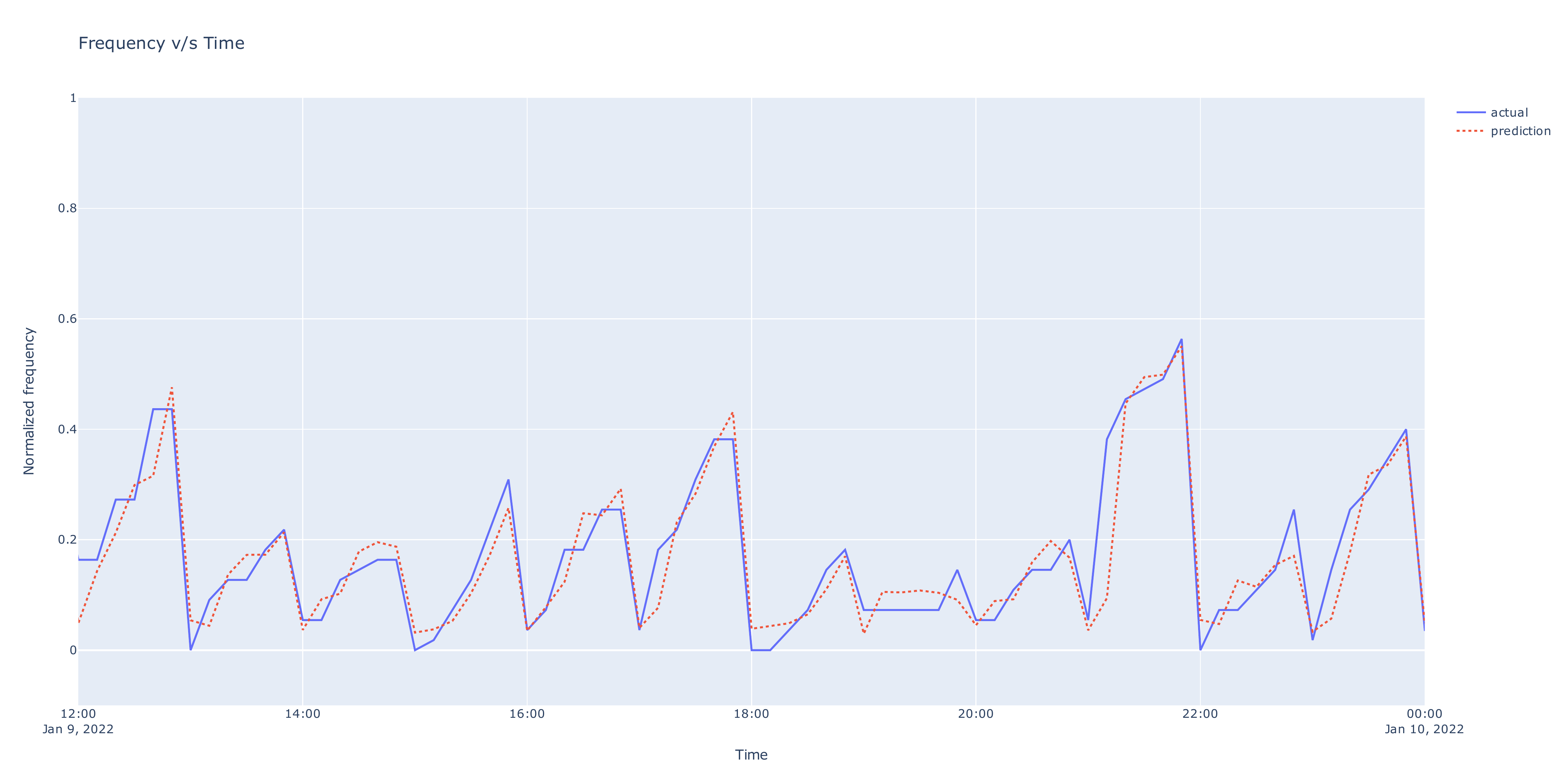}
		\caption{Lr=0.01, Ns=6}
		\label{fig-freq-vs-time-plot9}
	\end{subfigure}
	\caption{Impact of number of steps on predictions}
	\label{fig-freq-vs-time-plot7-8-9}
\end{figure}

Since for the majority of log analysis applications, a fast-learning model is required, in this work, the number of epochs and the learning rate are intentionally capped.
Therefore, although it is possible to further reduce the learning rate, this is not favorable as it requires a significantly larger number of epochs, which in turn extends the entire learning period.

Number of steps (Ns) also plays an equally important role in the model’s accuracy.
If too few or too many steps (e.g., 2 or 12) are considered, the model has almost no proper memory of the events.
Using these setups the model either projects the recently seen values as its predictions, or it predicts the overall trend in data.
This behavior is observed for all learning rates.
The best results as shown in Figure~\ref{fig-freq-vs-time-plot7} and Figure~\ref{fig-freq-vs-time-plot9} are achieved in 6 and 9 steps.
Although the prediction lags at certain points, in general, the model provides acceptable predictions.

\begin{figure}[!htb]
	\begin{subfigure}[b]{0.24\textwidth}
		\includegraphics[width=1\textwidth]{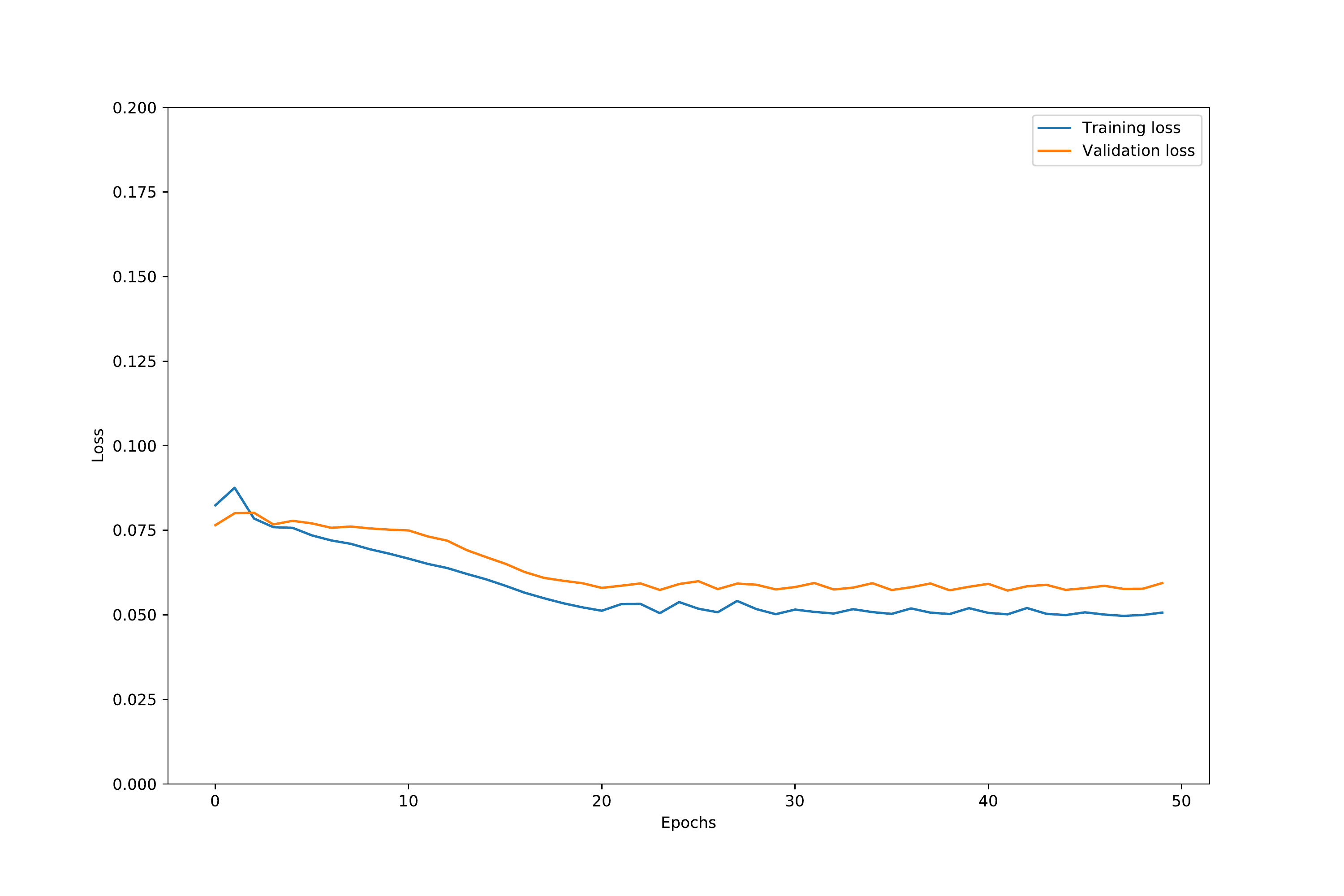}
		\caption{Lr=0.001,Ns=2}
		\label{fig-freq-vs-time-plot10_11}
	\end{subfigure}
	\begin{subfigure}[b]{0.24\textwidth}
		\includegraphics[width=1\textwidth]{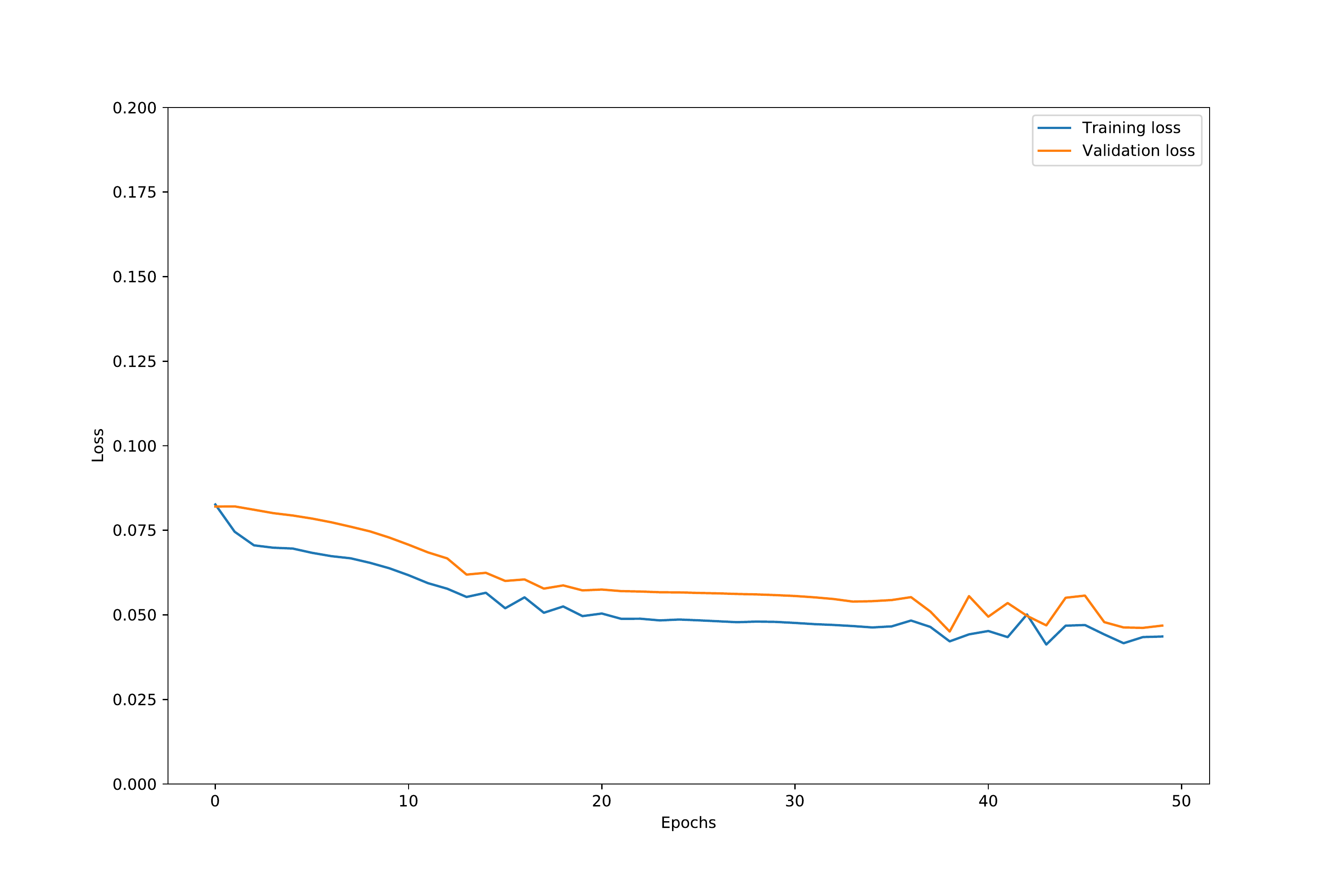}
		\caption{Lr=0.001,Ns=6}
		\label{fig-freq-vs-time-plot10_12}
	\end{subfigure}
	\begin{subfigure}[b]{0.24\textwidth}
	\includegraphics[width=1\textwidth]{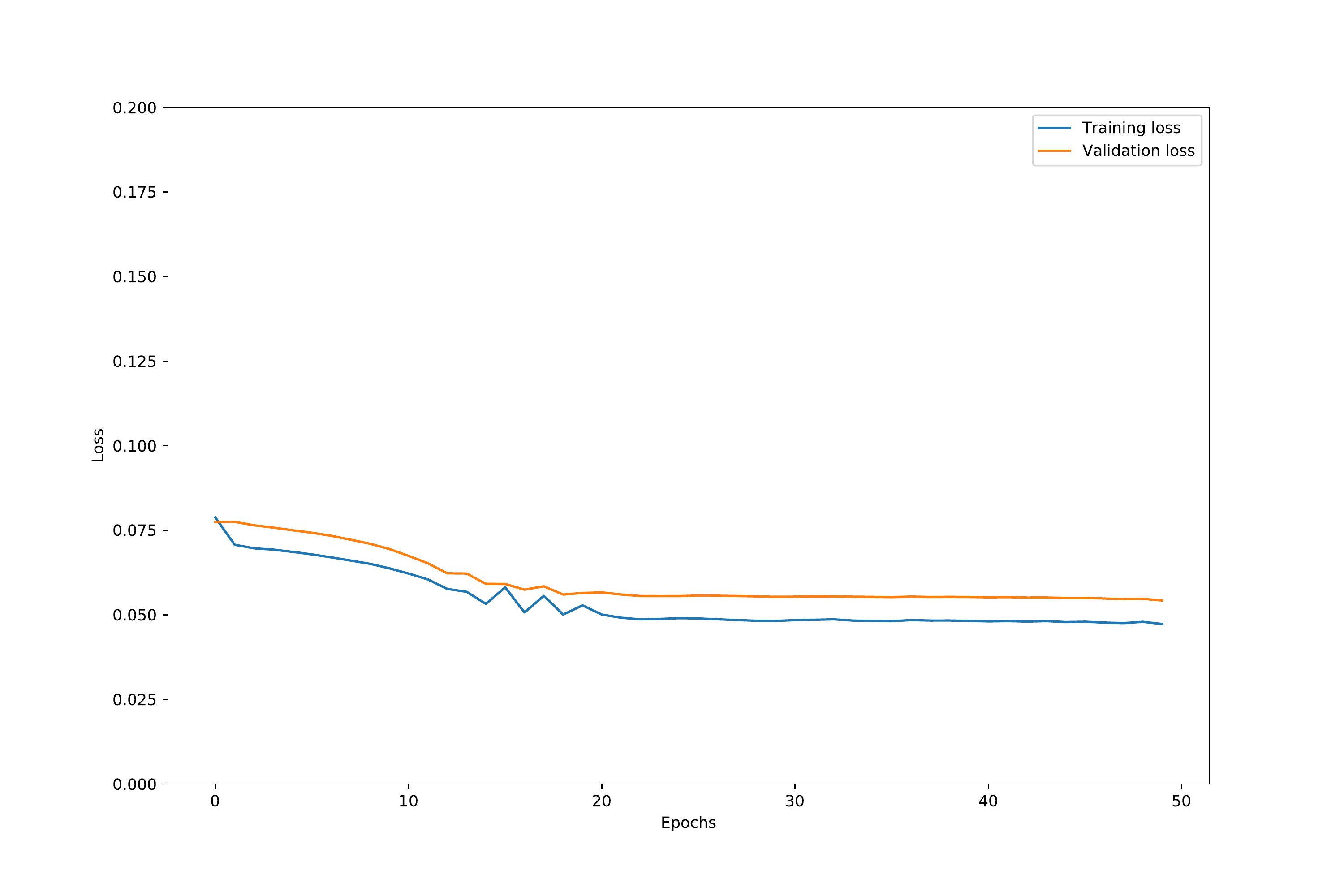}
	\caption{Lr=0.001,Ns=9}
	\label{fig-freq-vs-time-plot10_13}
\end{subfigure}
\begin{subfigure}[b]{0.24\textwidth}
	\includegraphics[width=1\textwidth]{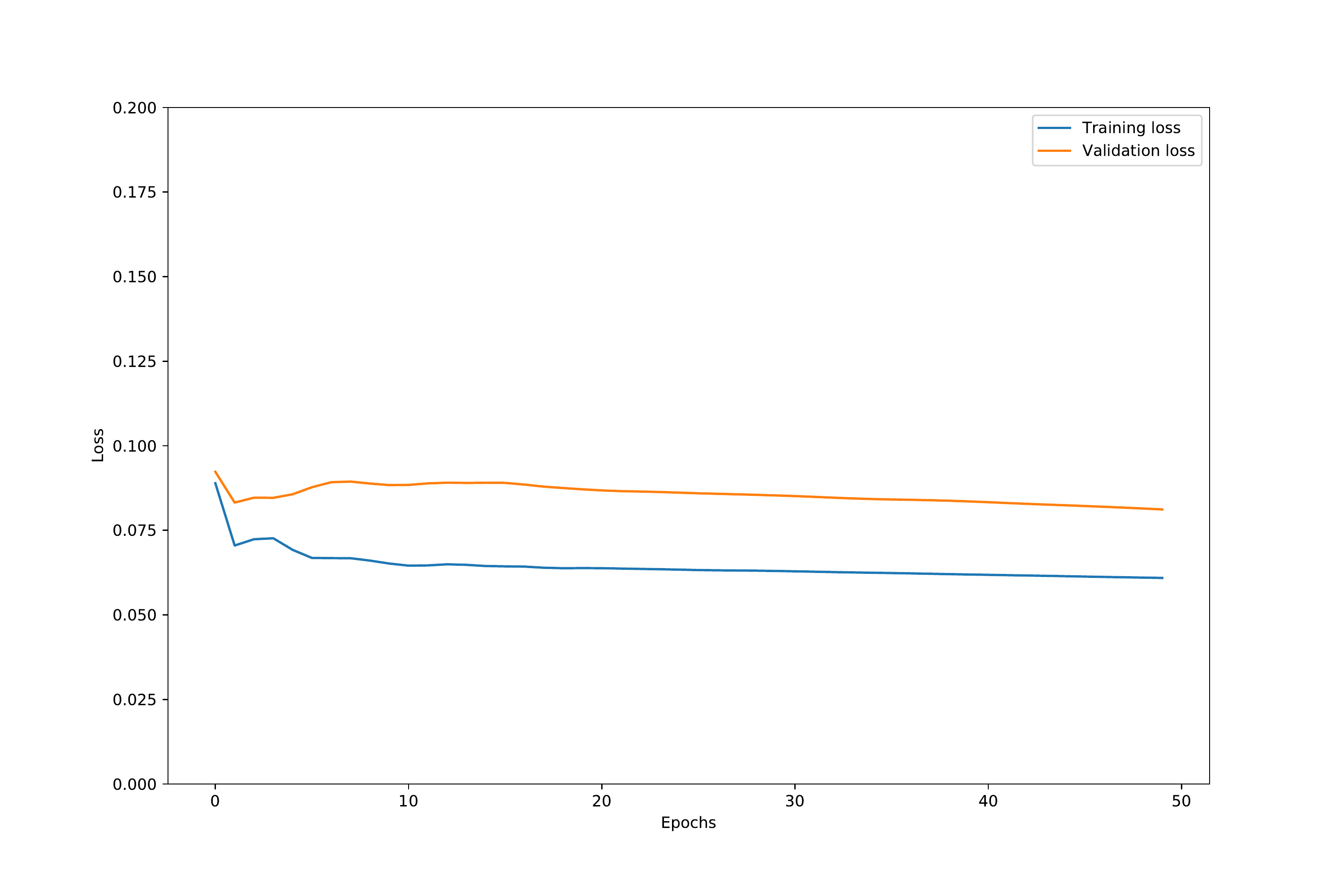}
	\caption{Lr=0.001,Ns=12}
	\label{fig-freq-vs-time-plot10_14}
\end{subfigure}
	\begin{subfigure}[b]{0.24\textwidth}
	\includegraphics[width=1\textwidth]{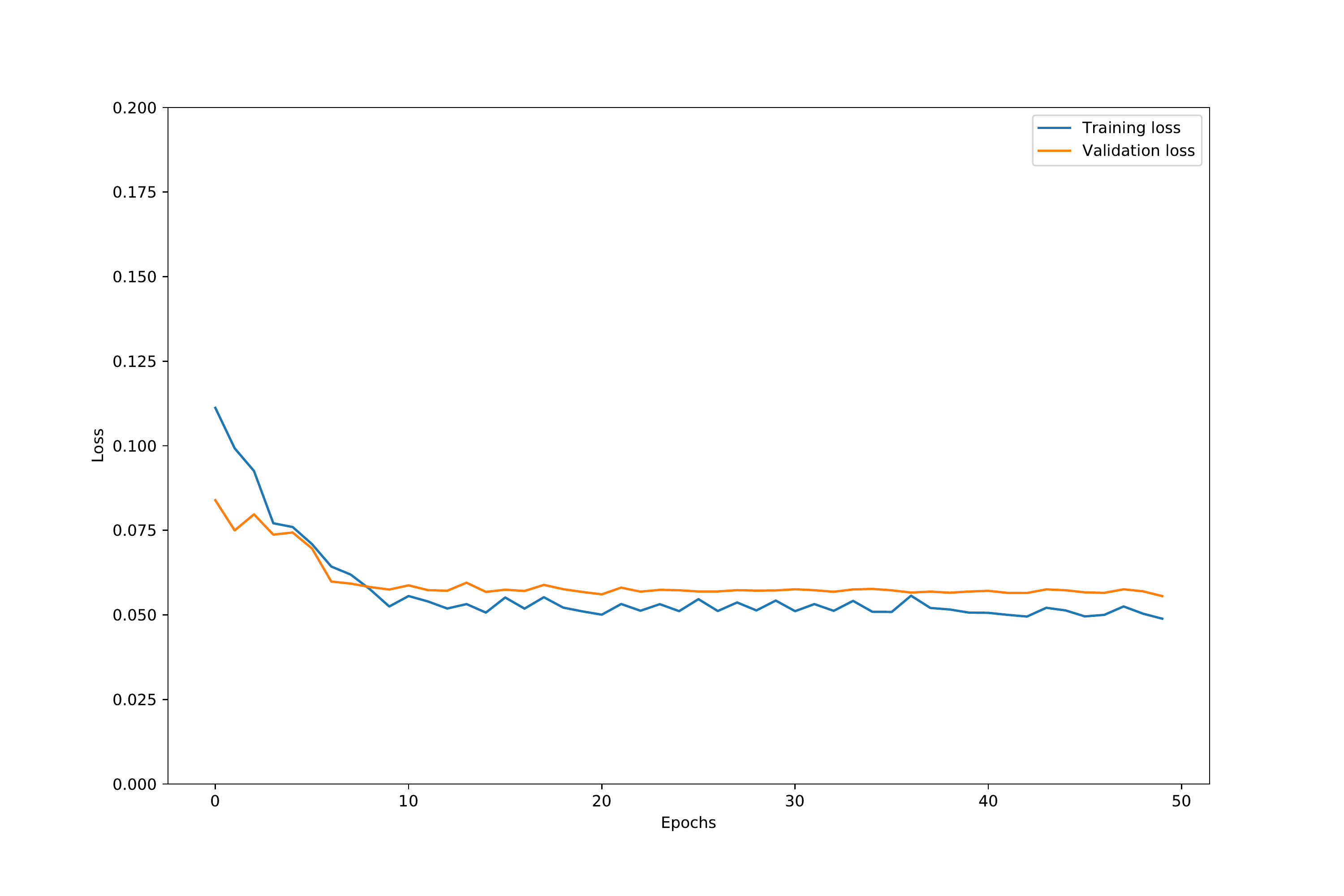}
	\caption{Lr=0.01,Ns=2}
	\label{fig-freq-vs-time-plot10_21}
\end{subfigure}
\begin{subfigure}[b]{0.24\textwidth}
	\includegraphics[width=1\textwidth]{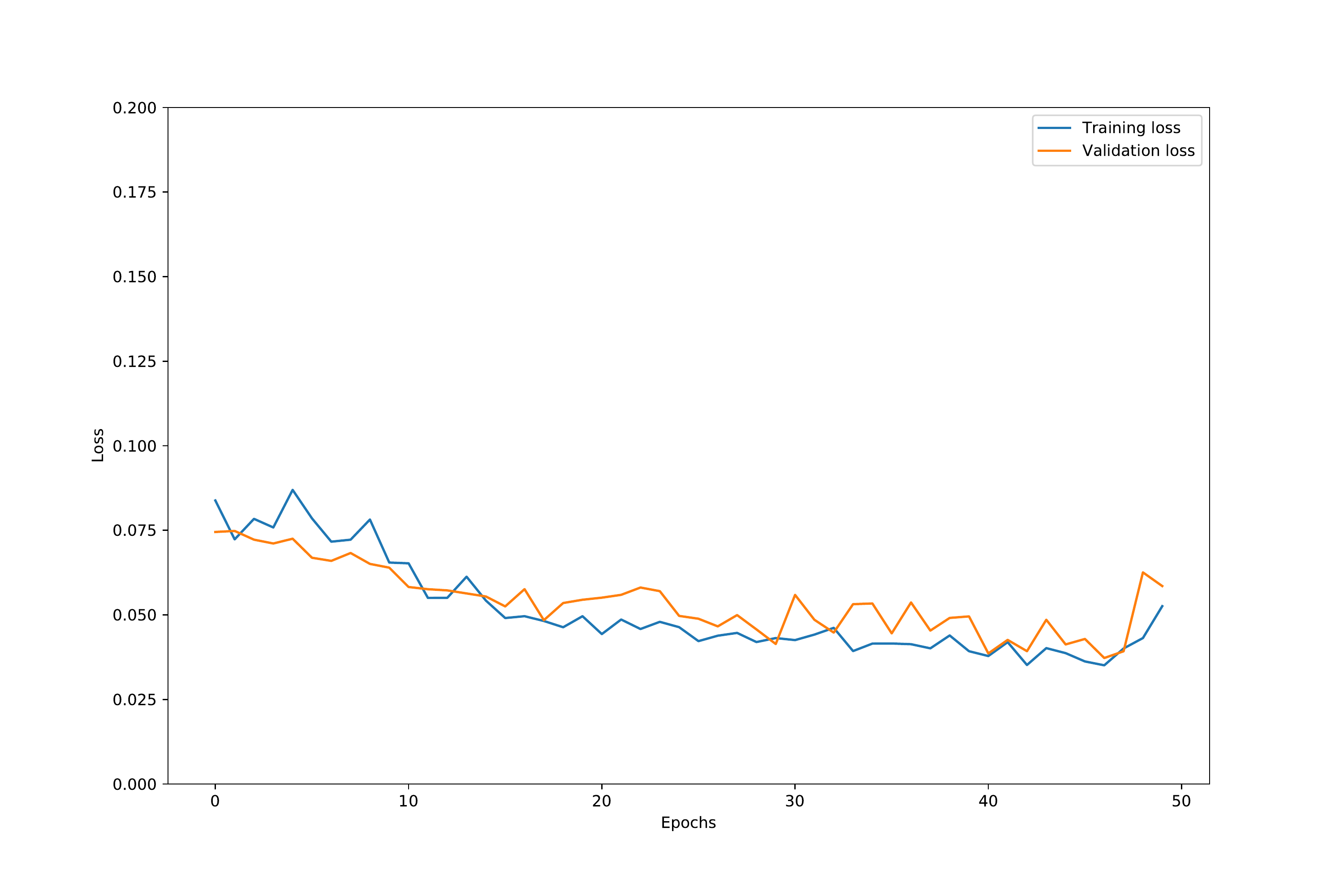}
	\caption{Lr=0.01,Ns=6}
	\label{fig-freq-vs-time-plot10_22}
\end{subfigure}
	\begin{subfigure}[b]{0.24\textwidth}
	\includegraphics[width=1\textwidth]{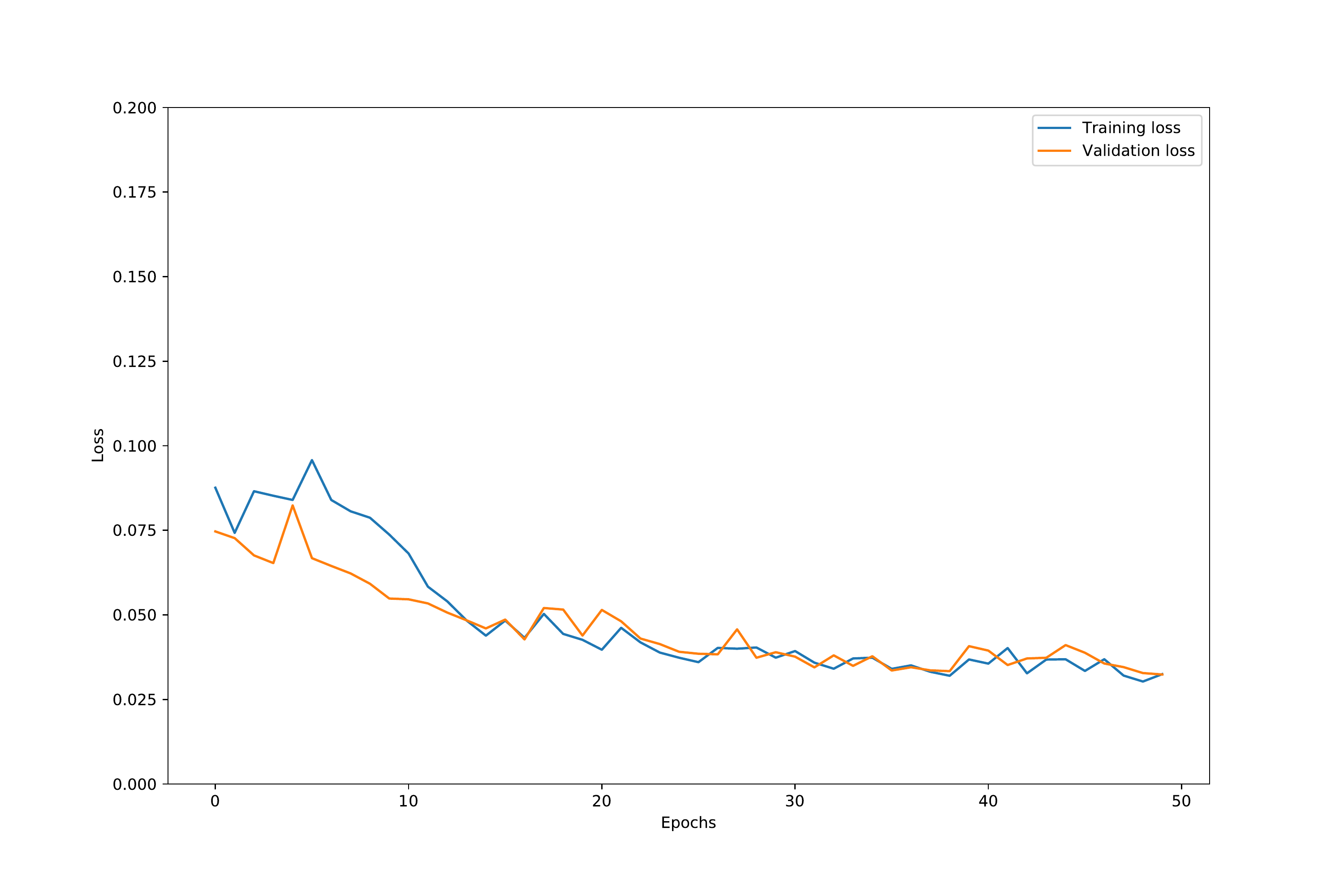}
	\caption{Lr=0.01,Ns=9}
	\label{fig-freq-vs-time-plot10_23}
\end{subfigure}
\begin{subfigure}[b]{0.24\textwidth}
	\includegraphics[width=1\textwidth]{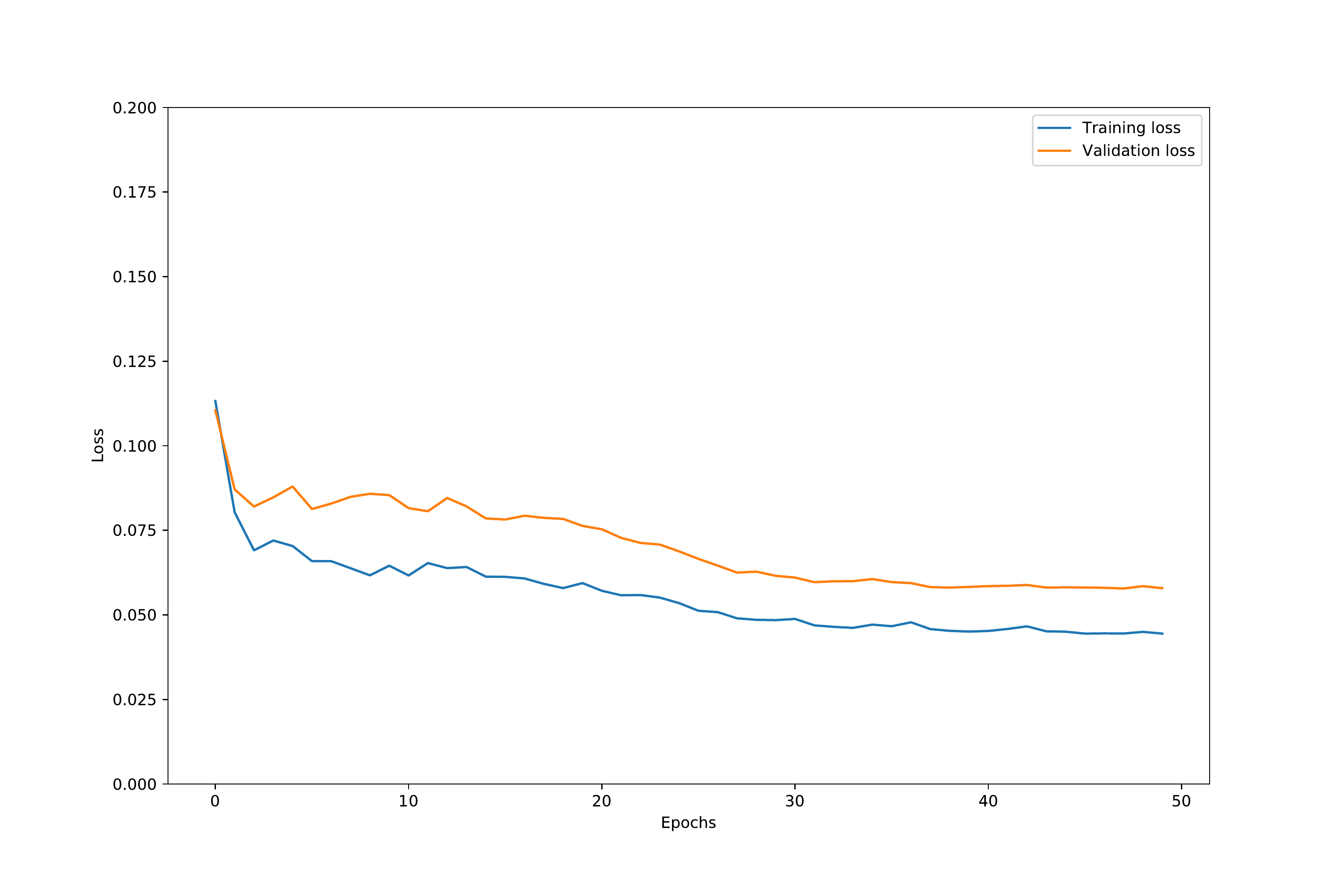}
	\caption{Lr=0.01,Ns=12}
	\label{fig-freq-vs-time-plot10_24}
\end{subfigure}
	\caption{Variation in loss plot for 10 min bucket with changing learning rates and and number of steps}
	\label{fig-freq-vs-time-plot10}
\end{figure}

Traces of routine system operations that typically occur in one-hour cycles are recorded in system logs.
Therefore, observing a more accurate prediction based on the last 6 steps (60 minutes) was not unexpected.
Similar observations were made for the 30-minute time bucket.
Compared to the coarser time buckets, the 10-minute bucket provides better prediction accuracy.
Fig.~\ref{fig-freq-vs-time-plot10} illustrates an overview of the final results.

\paragraph{Multivariate data}
The multivariate setup requires different parameter combinations in comparison to the univariate setup.
Therefore, similar experiments were done to evaluate the usefulness of anonymized system logs for multivariate analysis via the LSTM model.
The outcome of multivariate analysis using the provided parameters is shown in Table~\ref{tab-training-validation-uni-multi}.

As expected, using 6-time steps provides better results compared to using only 2 time steps.
This is intuitive as with 2-steps the model has limited memory and has a lagging prediction.
Although the training and validation loss gets slightly better with different adjustments, none of these setups result in a sufficiently accurate model as seen for the univariate dataset.

\paragraph{Automatic hyperparameter optimization:}
To further optimize the set of hyperparameters for the model, additional testing was done on the univariate data using the Keras Tuner\footnote{Available at https://keras.io/keras\_tuner/} library.
According to Figure~\ref{fig-freq-vs-time-plot7}, the bucket size of 10 minutes and 9 steps were chosen for this experiment.
The same model as defined in Section~\ref{subsec-lstm-model} was employed.
The learning rate was varying between 0.0001 and 0.001 and the mean squared logarithmic error (MSLE) was chosen as the loss function.
It is worth mentioning that MSLE was used for the Keras tuner as using other loss functions resulted in the model getting stuck in a local minimum.

\begin{figure}[!htb]
	\begin{subfigure}[b]{0.24\textwidth}
		\includegraphics[width=1\textwidth]{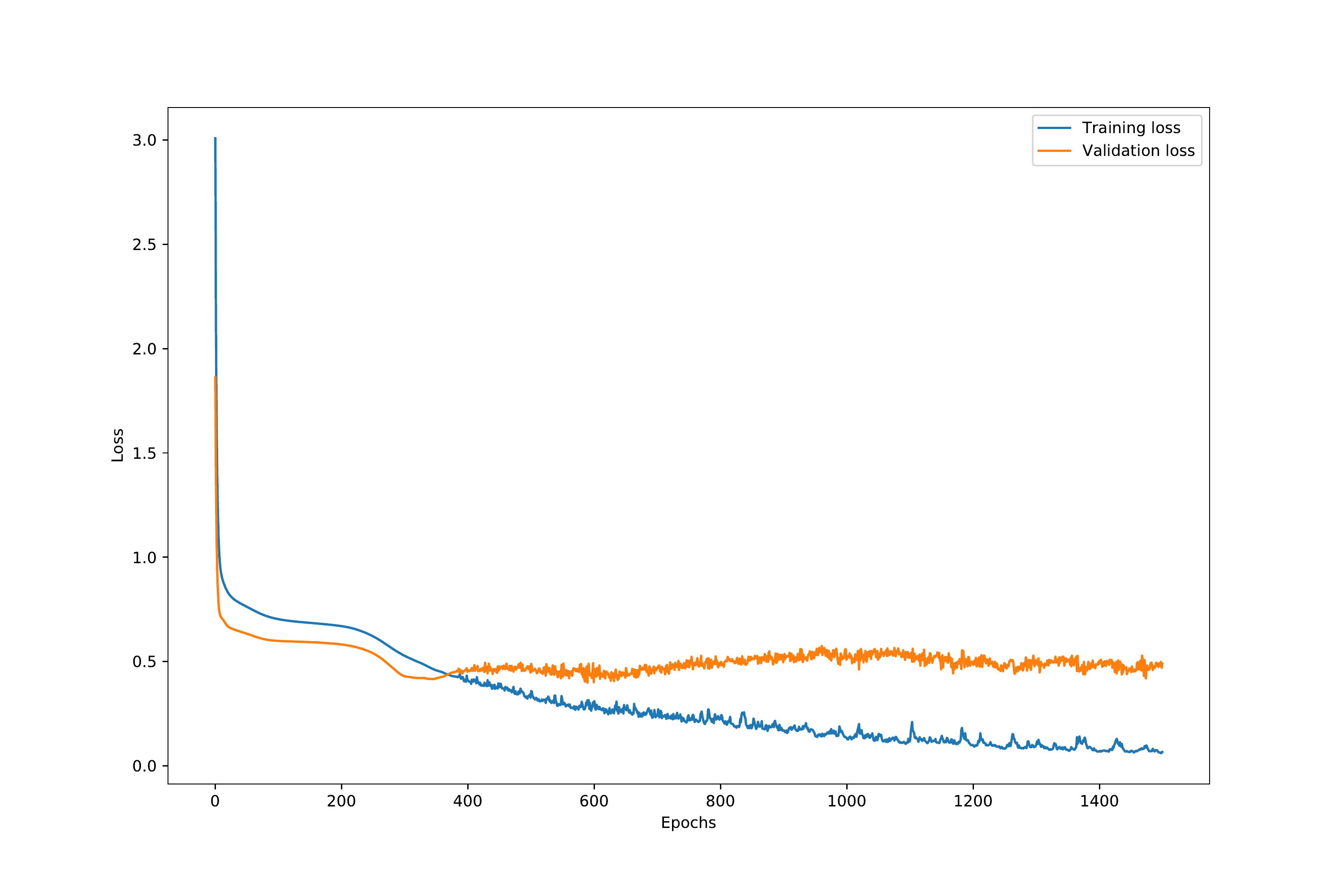}
		\caption{Lr=0.0001, 1-L}
		\label{fig-loss-1e-4_layer1}
	\end{subfigure}
	\begin{subfigure}[b]{0.24\textwidth}
		\includegraphics[width=1\textwidth]{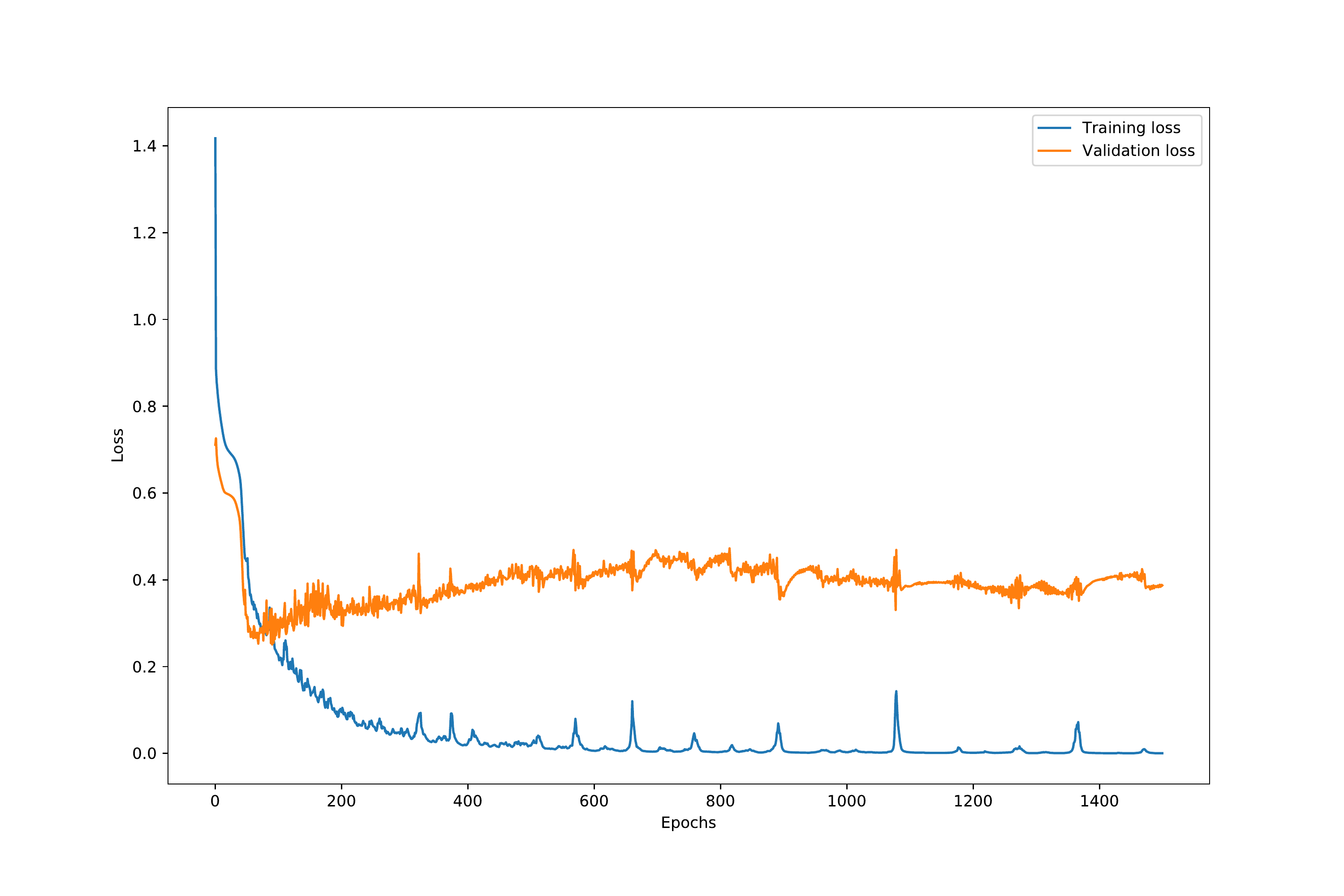}
		\caption{Lr=0.001, 1-L}
		\label{fig-loss-1e-3_layer1}
	\end{subfigure}
	\begin{subfigure}[b]{0.24\textwidth}
		\includegraphics[width=1\textwidth]{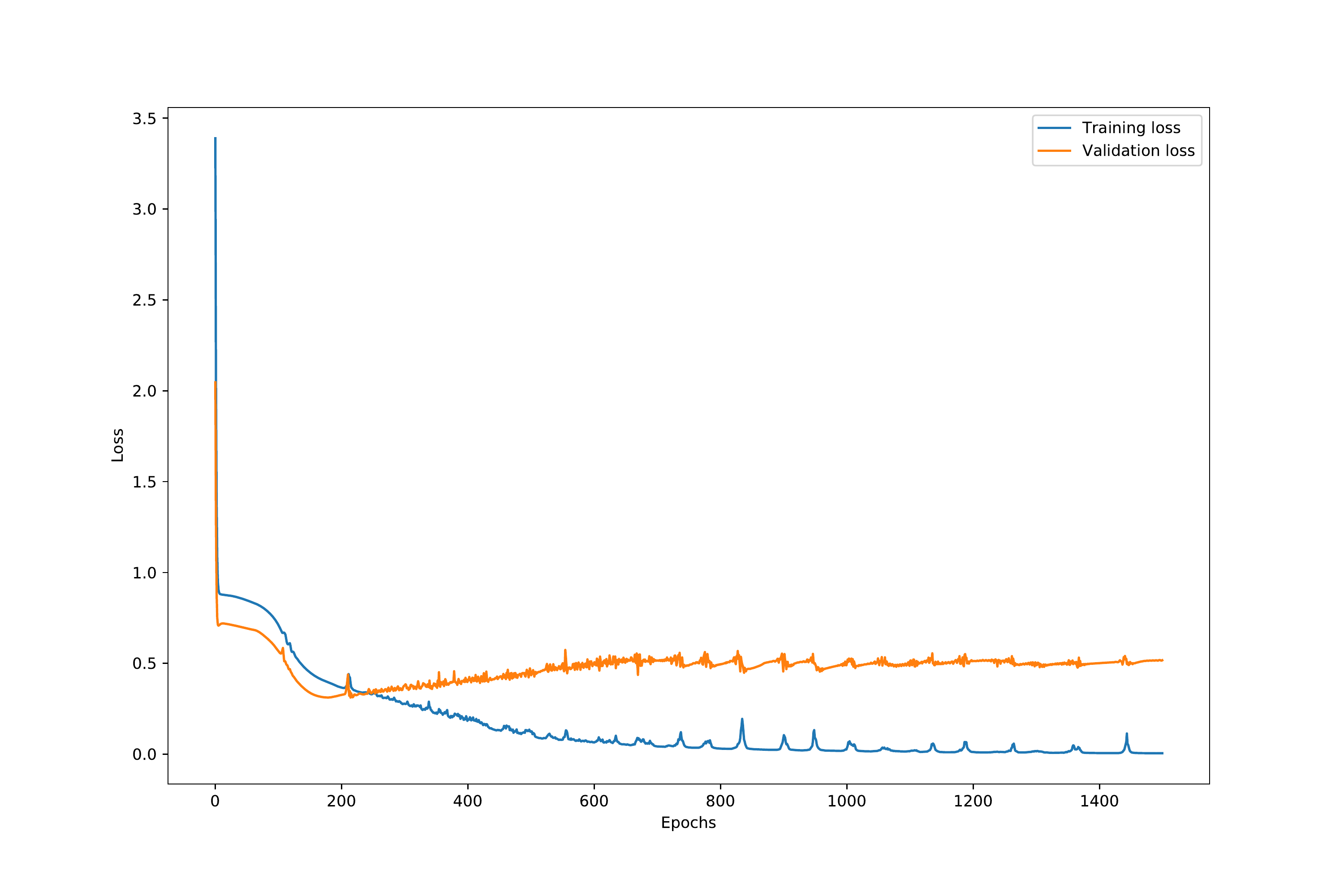}
		\caption{Lr=0.0001, 3-L}
		\label{fig-loss-1e-4_layer3}
	\end{subfigure}
	\begin{subfigure}[b]{0.24\textwidth}
		\includegraphics[width=1\textwidth]{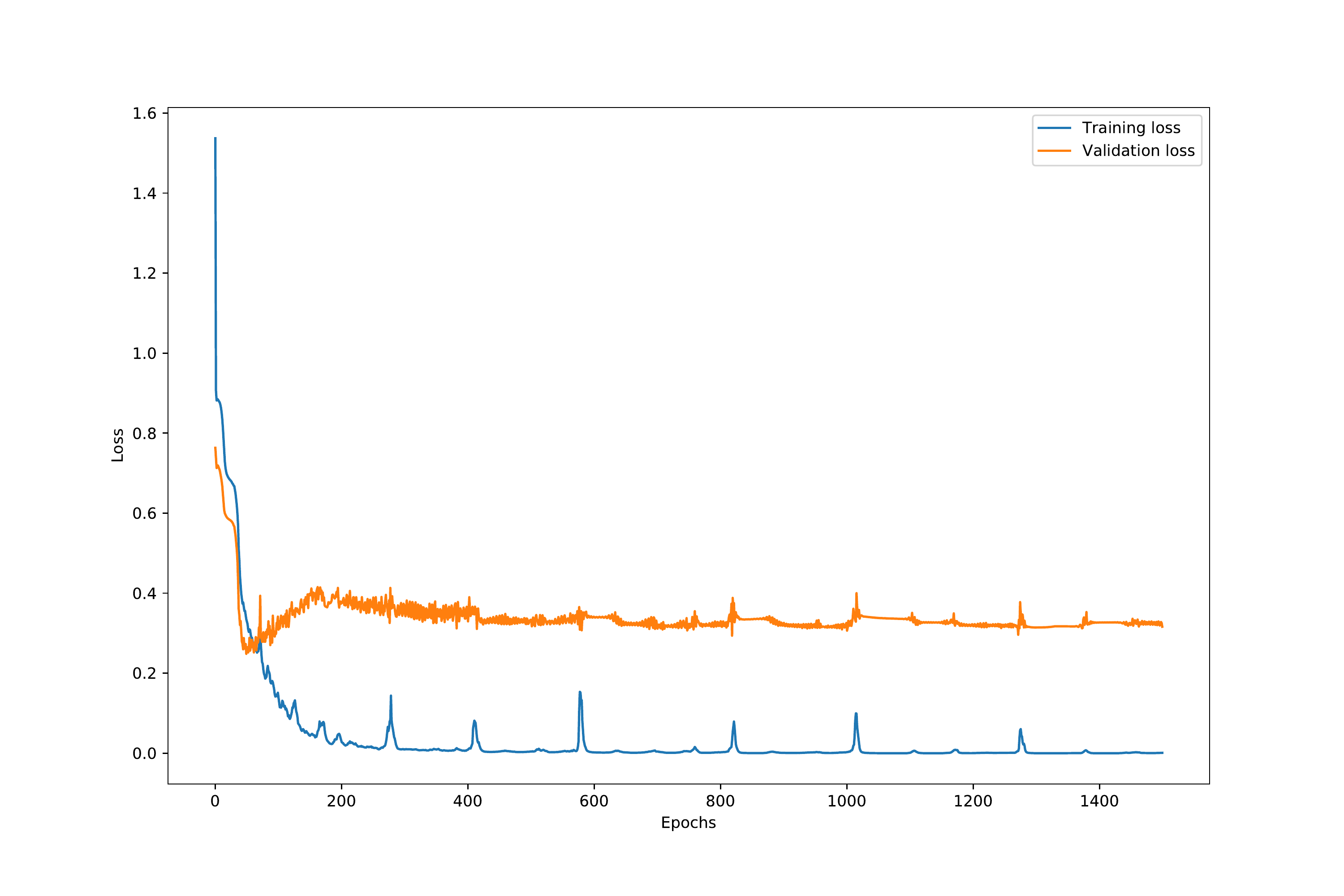}
		\caption{Lr=0.001, 3-L}
		\label{fig-loss-1e-3_layer3}
	\end{subfigure}
	\caption{Loss plots for 10 min bucket and 9 steps over 1500 epochs.}
	\label{fig-loss-kt}
\end{figure}

In addition, to explore the impact of model complexity, the number of LSTM layers in the proposed model was alternating between 1 and 3 layers.
To assure the correctness and completeness of the observations, the number of epochs was also increased to 1500.
Figure~\ref{fig-loss-kt} provides an overview of the outcome in various setups.

The best model with 128 neurons and a learning rate of 0.0006 selected by the Keras tuner did not provide any significant improvement compared to the initial setup, built by manual experiments.
Furthermore, additional layers of LSTM did not improve the predictions.
In a single-layer setup, overfitting of the model after 100 epochs confirms the correctness of choosing less number of epochs for the provided dataset.

An interesting observation made from Figure~\ref{fig-loss-1e-3_layer1} is the presence of a mirroring effect on training and validation loss.
This could be an indication of noises in the monitoring data.
Thus, further pre-processing steps might be necessary to improve the accuracy of models trained by anonymized system logs.

Finally, it can be concluded that univariate analysis is more effective for training LSTM models using anonymized system logs encoded via the $P\alpha RS$ anonymization method.
Despite the improvements observed for varying setups, generally, the effect of training is missing which can be seen from most of the loss plots. This could be a consequence of the anonymization done on the data. However, the usefulness of this data for anomaly detection can still be confirmed from this analysis.

\section{Conclusion and future works}
\label{sec-conclusion}
This work provided a comprehensive assessment on the usefulness of fully anonymized monitoring data for anomaly detection using LSTM models.
System logs due to their availability on current HPC systems and their information richness are desired monitoring data for behavioral analysis.
In addition, conclusions derived from Syslog analysis can be generalized to a wide range of computing systems.
To address the privacy concerns raised due to the existence of sensitive data in system logs, the $P\alpha RS$ anonymization mechanism is applied.
$P\alpha RS$ preserves the similarity of system logs while encoding them into a stream of hashed messages.   
Based on previous works, it was known that 80\% of the system logs on Taurus are generated by the top 10 most frequent patterns.
The normal behavior of the system is expected to be dictated by these highly frequent patterns.
Therefore, the frequency of appearance of the top 10 patterns among resulting anonymized system logs, calculated within a specified time bucket, was chosen as the quantitative metric.
An unsupervised machine learning model, namely LSTM was chosen.
The size of the model, the amount of required data, and the number of epochs were kept to their minimum, to match the dynamic nature of HPC monitoring data.
The selection of effective hyper-parameters was made rationally by considering the prior information.
The prototype was implemented in Python using Keras.
The python code, Syslog data, and information on how to reproduce this work are available at~\cite{ref_gitcode}.

According to the analysis, the system logs anonymized by $P\alpha RS$, despite the complete concealment of log information, are still usable even in the simplest LSTM models for behavioral analysis.
Therefore, the usefulness of such anonymized data for anomaly detection via LSTMs is confirmed.
However, the best model found after fine-tuning was seen to predict the pattern to a certain extent but not with significantly high accuracy, mainly caused by the shortcomings present in the monitoring data.
In future work, more quantitative data such as power consumption and temperature variations will be added to the model, to further improve the accuracy of behavioral analysis.
Significant deviation from the identified normal systems behavior could be the signal of potential anomalous behavior.
However, defining the correct criteria for such a threshold is a challenging topic which in future works will be addressed.
Furthermore, implementing a robust pipeline for anomaly detection is planned.

\begin{acknowledgments}
	This work was supported by the German Federal Ministry of Education and Research (BMBF, 01/S18026A-F) by funding the competence center for Big Data and AI "ScaDS.AI Dresden/Leipzig".
	The authors gratefully acknowledge the GWK support for funding this project by providing computing time through the Center for Information Services and HPC (ZIH) at TU Dresden on HRSK-II. 
\end{acknowledgments}

%
%
%
%

\end{document}